\documentclass[10pt,twocolumn,letterpaper]{article}
\usepackage{microtype}
\usepackage{graphicx}
\usepackage{subfigure}
\usepackage{multirow}
\usepackage{multirow}   
\usepackage{booktabs}   
\usepackage{bm}         
\usepackage{array}
\usepackage{booktabs}
\usepackage{rotating}
\usepackage{graphicx}
\usepackage{algorithm}
\usepackage{algpseudocode}
\usepackage{amsmath}
\usepackage[table]{xcolor} 

\definecolor{grey}{HTML}{D9D9D9}

\newcommand*\rot{\rotatebox{70 }}

\usepackage{amsmath}
\usepackage{amssymb}
\usepackage{mathtools}
\usepackage{amsthm}

\usepackage{iccv}              

\definecolor{iccvblue}{rgb}{0.21,0.49,0.74}
\usepackage[pagebackref,breaklinks,colorlinks,allcolors=iccvblue]{hyperref}

\def\paperID{12188} 
\def\confName{ICCV}
\def\confYear{2025}

\title{\includegraphics[width=6mm]{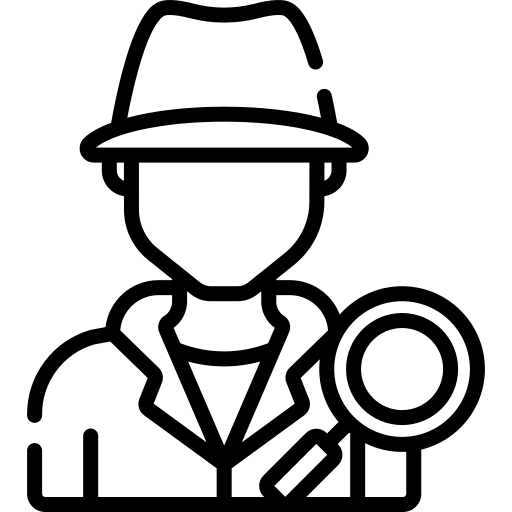} \textit{AIGI-Holmes}: Towards Explainable and Generalizable AI-Generated Image Detection via Multimodal Large Language Models}

\author{
Ziyin Zhou$^{1*\#}$\quad
Yunpeng Luo$^{2*}$\quad
Yuanchen Wu$^{2}$\quad
Ke Sun$^{1}$\quad 
Jiayi Ji$^{1}$\quad \\
Ke Yan$^{2{\dag}}$\quad 
Shouhong Ding$^{2}$\quad
Xiaoshuai Sun$^{1{\dag}}$\quad
Yunsheng Wu$^{2}$\quad
Rongrong Ji$^{1}$\quad
\\
$^{1}$ Key Laboratory of Multimedia Trusted Perception and Efficient Computing, \\ \quad \quad Ministry of Education of China, Xiamen University\quad
$^{2}$ Tencent YouTu Lab \\
{\tt\scriptsize \{ziyinzhou,skjack\}@stu.xmu.edu.cn\quad \{xssun,rrji\}@xmu.edu.cn\quad } \\ {\tt\scriptsize jjyxmu@gmail.com} {\tt\scriptsize \{petterluo,kerwinyan\}@tencent.com} \\ {\tt\small  \url{{https://github.com/wyczzy/AIGI-Holmes}}}
} 

\begin{document}
\maketitle

\begin{abstract}
The rapid development of AI-generated content (AIGC) technology has led to the misuse of highly realistic AI-generated images (AIGI) in spreading misinformation, posing a threat to public information security. Although existing AIGI detection techniques are generally effective, they face two issues: 1) a lack of human-verifiable explanations, and 2) a lack of generalization in the latest generation technology. To address these issues, we introduce a large-scale and comprehensive dataset, Holmes-Set, which includes the Holmes-SFTSet, an instruction-tuning dataset with explanations on whether images are AI-generated, and the Holmes-DPOSet, a human-aligned preference dataset. Our work introduces an efficient data annotation method called the Multi-Expert Jury, enhancing data generation through structured MLLM explanations and quality control via cross-model evaluation, expert defect filtering, and human preference modification. In addition, we propose Holmes Pipeline, a meticulously designed three-stage training framework comprising visual expert pre-training, supervised fine-tuning, and direct preference optimization. Holmes Pipeline adapts multimodal large language models (MLLMs) for AIGI detection while generating human-verifiable and human-aligned explanations, ultimately yielding our model AIGI-Holmes. During the inference stage, we introduce a collaborative decoding strategy that integrates the model perception of the visual expert with the semantic reasoning of MLLMs, further enhancing the generalization capabilities. Extensive experiments on three benchmarks validate the effectiveness of our AIGI-Holmes.
\end{abstract}    
\section{Introduction}
The rapid evolution of AI technologies like GANs~\cite{goodfellow2014generative, karras2019style} and Diffusion models~\cite{dhariwal2021diffusion, rombach2022high} has made generated images highly realistic. While beneficial for digital art and film, these technologies also pose risks such as misinformation, privacy breaches, and deepfakes. Recent advancements in diffusion models (e.g., FLUX~\cite{flux}, SD3, SD3.5~\cite{esser2024scaling}) and autoregressive techniques (e.g., VAR~\cite{tian2024visual}) have further complicated detection, highlighting the urgent need for effective AI-generated image detection methods.

\renewcommand{\thefootnote}{}
\footnotetext{\#~This work was done during an internship at Tencent YouTu Lab.}
\footnotetext{*~Equal Contribution.}
\footnotetext{\dag~Corresponding Author.}
\renewcommand{\thefootnote}{\arabic{footnote}}
\begin{figure*}[!htbp]
\centering
\vspace{-0.em}
\includegraphics[width=0.99\textwidth]{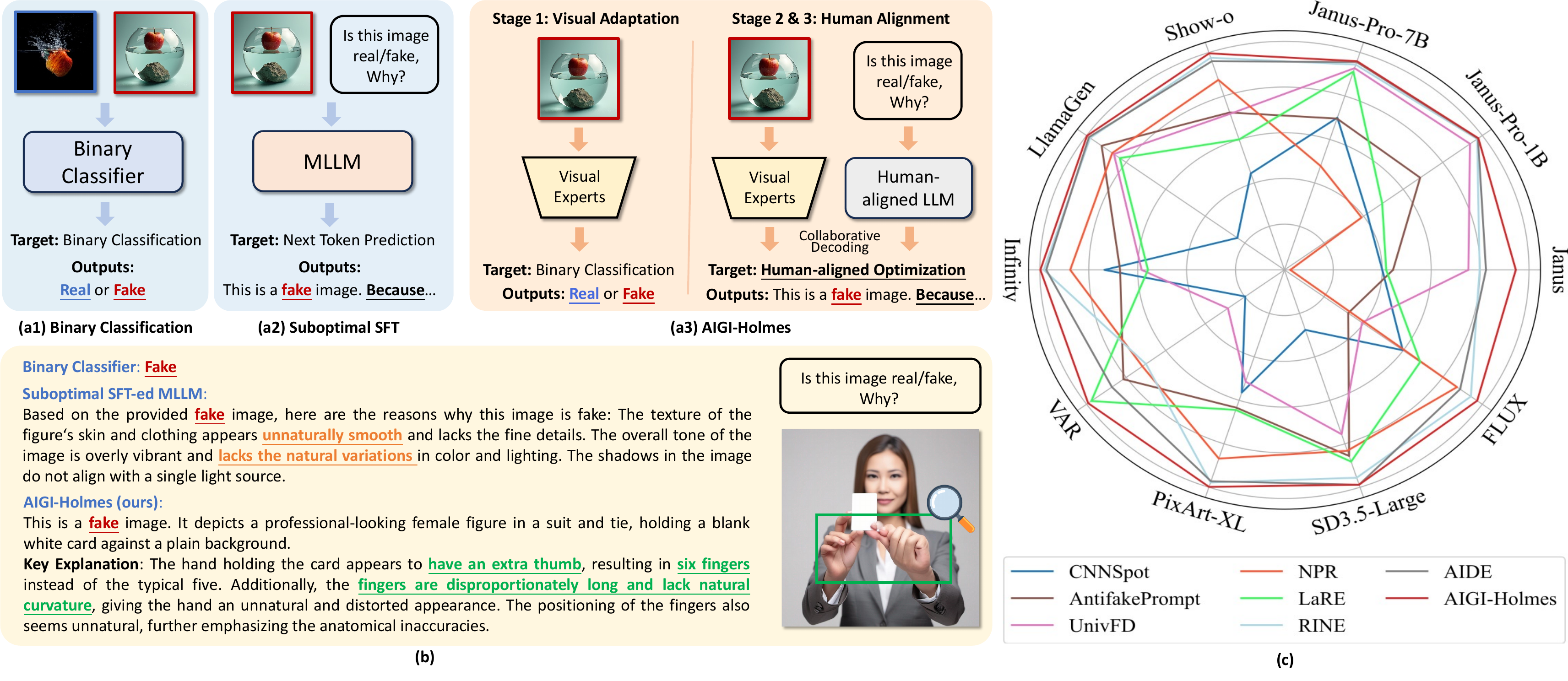}
\vspace{-1.1em}
\caption{(a): Comparison of AIGI-Holmes with existing methods, (b): A qualitative example to illustrate the effect of AIGI-Holmes, (c): AIGI-Holmes outperforms existing baseline methods on state-of-the-art generators under unseen settings.} 
\label{fig:intro}
\vspace{-1.5em}
\end{figure*}

Recent studies~\cite{wang2020cnn,wang2023dire, tan2024rethinking, luo2024lare, koutlis2024leveraging, yan2024sanity} have demonstrated remarkable advancements in AI-generated image detection. While these improvements are noteworthy, two critical issues limit their application and generalization in real-world scenarios: \textbf{1. Lack of explanation:} Current detection models are black boxes (Fig.~\ref{fig:intro}(a) , making their detection results difficult for humans to verify. The lack of human-verifiable explanations leads to unreliable detection results. \textbf{2. Lack of generalization:} The rapidly evolving AIGC technologies (Fig.~\ref{fig:intro}(c)) persistently challenge their generalization capabilities. Therefore, developing explainable and generalizable AI-generated image detection algorithms is becoming increasingly urgent.

Recent breakthroughs in Multimodal Large Language Models (MLLMs)~\cite{zhang2025common, huang2024ffaa, chen2024textitx2dfdframeworkexplainableextendable, lian2024large, xu2024fakeshield, sun2024forgerysleuth, huang2024sida, yang2024heie, qu2024explainable} offer a promising pathway: their exceptional capabilities in commonsense understanding and natural language generation enable semantic-level analysis of visual content. This makes MLLMs a strong candidate for explainable and generalizable AI-generated image detection algorithms. There have been preliminary explorations of using MLLMs for AI-generated images~\cite{chang2023antifakeprompt, keita2024bi}. However, merely employing MLLMs for binary classification predictions does not fully leverage their potential. We aspire for Multimodal Large Language Models (MLLMs) to emulate the capabilities of the renowned detective Sherlock Holmes, not only by accurately identifying the culprit but also by providing precise and corresponding evidence, thereby offering human-verifiable explanations.

However, two key challenges hinder MLLM's effective use in AI-generated image detection: 1). \textbf{Training Data Scarcity}: As shown in Tab.~\ref{tab:data_annotation1}, existing AI-generated image detection datasets, such as CNNDetection~\cite{wang2019cnngenerated}, GenImage~\cite{zhu2023genimage}, and DRCT~\cite{pmlr-v235-chen24ay}, consist solely of visual modalities and lack instruction fine-tuning datasets suitable for the SFT(Supervised Fine-tuning) phase of MLLMs. Although FakeBench~\cite{li2024fakebench} and LOKI~\cite{ye2024loki} make initial attempts in this field, the use of GPT-4o~\cite{achiam2023gpt} and the high cost of human labor limit the expansion of these datasets, rendering them unsuitable for SFT of MLLMs. 2). \textbf{Suboptimal Supervised Fine-Tuning.} Simply training MLLMs on Supervised Fine-Tuning datasets (even with explanations) often leads to sub-optimal performance (as shown in Tab.~\ref{table:exp} and Tab.~\ref{tab_abl}). Possible reasons include: 1. MLLMs exhibit somewhat insufficient capabilities in image classification tasks~\cite{zhang2024visually} and low-level perception~\cite{wu2023q}, which are often closely related to the generalization of AI-generated image detection. 2. SFT models may mechanically replicate explanation templates without genuinely understanding the underlying causes of artifacts or semantic errors. 

To address the issues with the training dataset, we first introduce two key datasets: Holmes-SFTSet and Holmes-DPOSet. The Holmes-SFTSet provides 65K images annotated with explanations across high-level semantic dimensions (e.g., physical inconsistencies, anatomical errors, and text rendering flaws) and low-level artifacts (e.g., overall hue, texture, edges), rigorously refined through cross-model validation and expert-guided filtering to ensure alignment with human-verifiable evidence. To further the critical need for human-aligned judgment in AI-generated image detection, we introduce the Holmes-DPOSet by constructing contrastive explanation pairs through Positive/Negative prompts on 65K images from the Holmes-SFTSet, along with iterative expert corrections for an additional 4K image explanations, effectively bridging the gap between model perception and human reasoning.

Building upon this training dataset, we propose the Holmes Pipeline, which introduces three key stages to enhance the generalization and interpretability of AIGI detection through systematic MLLM fine-tuning to address the issue of Suboptimal Fine-tuning. The process begins with Visual Expert Pre-training, which leverages the Holmes-SFTset to rapidly adapt the visual encoder through binary classification, establishing domain-specific feature extraction. Following this, Supervised Fine-Tuning enables MLLMs to not only detect synthetic content but also generate human-verifiable explanations. This stage addresses the ``black box'' limitation of conventional binary classification approaches. Finally, Human-aligned Direct Preference Optimization~\cite{rafailov2024direct} utilizes the Holmes-DPOset. This stage fundamentally reshapes the reasoning patterns of MLLMs by learning from preference samples, ensuring that the interpretative results align with human judgment standards rather than suboptimal fine-tuning. During inference, our collaborative decoding strategy integrates the model perception of the visual expert with the semantic reasoning of MLLMs, creating a dual-channel verification process that enhances the generalizability of our approach. Our method demonstrates superior performance over state-of-the-art approaches in AI-generated image detection, while producing human-aligned explanations that enhance detection reliability.

In summary, our contributions are threefold:
\begin{itemize}
\item \textbf{Dataset:} We introduce Holmes-SFTSet and Holmes-DPOSet, the first explanation-rich datasets that include human-verifiable evidence through semantic annotations and contrastive preference pairs, addressing the critical training data scarcity in AI-generated image detection.
\item \textbf{Methodology:} We propose the Holmes Pipeline, a systematic training pipeline for multimodal large language models (MLLMs) that includes visual expert pre-training, explanation-aware supervised fine-tuning, and human-aligned direct preference optimization. This pipeline synergizes model perception with semantic reasoning through novel collaborative decoding during the inference process.
\item \textbf{Performance:} Our method achieves state-of-the-art detection accuracy on three benchmark datasets and provides human-verifiable explanations, demonstrating superior generalizability and alignment with human judgment.
\end{itemize}

\section{Related Work}
\label{data}
\subsection{Detection of AI-Generated Fake Images}
With the advancement of AI-based image generation technologies, numerous detection methods have emerged, focusing on training on a single AI-generated image method and generalizing to a wide range of AI-generated images. CNNSpot~\cite{wang2020cnn} finds that classifiers trained on ProGAN~\cite{karras2017progressive} can generalize to unseen GANs using data augmentation. FreDect~\cite{frank2020leveraging} detects anomalies in the frequency domain of GAN-generated images. Recent methods have explored new perspectives for better generalization. UnivFD~\cite{ojha2023towards} proposes to use pre-trained CLIP-ViT~\cite{radford2021learning} features, generalizing to out-of-distribution (OOD) data through nearest neighbor and linear probing. DIRE~\cite{wang2023dire} introduces Diffusion Reconstruction Error (DIRE), distinguishing real images from Diffusion Model (DM)-generated images by measuring reconstruction errors. DRCT~\cite{chendrct} introduces Reconstruction Contrastive Learning (RCL), enhancing generalization by generating challenging samples. PatchCraft~\cite{zhong2024patchcraft} detects generated images by segmenting them into small patches, applying SRM filters~\cite{fridrich2012rich}, and examining pixel correlations. NPR~\cite{tan2024rethinking} introduces neighboring pixel relationships, identifying generated content by analyzing local pixel distribution patterns during upsampling. AIDE~\cite{yan2024sanity} develops a two-stream framework using both frequency and semantic information. Despite their acceptable performance, these detection methods cannot explain their underlying principles and struggle to generalize against advanced AI-generated techniques.

\subsection{Multimodal Large Language Models}
Recent advancements in multimodal language large models (MLLMs) have enhanced Image Forgery Detection and DeepFake Detection, enabling interpretable methods. In DeepFake Detection, DD-VQA~\cite{zhang2025common} and FFAA~\cite{huang2024ffaa} pioneer the use of MLLMs, leveraging human and GPT-4o~\cite{hurst2024gpt} annotated datasets to train models based on InstructBLIP~\cite{instructblip} and LLaVA~\cite{liu2024visual}. In Image Forgery Detection, FAKESHIELD~\cite{xu2024fakeshield} combines MLLMs with a visual segmentation model (SAM~\cite{kirillov2023segment}) for explainable detection (e-IFDL), using a GPT-4o-enhanced dataset (MMTD-Set). ForgerySleuth~\cite{sun2024forgerysleuth} uses MLLMs with a trace encoder to detect tampering and generate detailed analyses, creating the ForgeryAnalysis dataset. For Text Tampering Detection, TextSleuth~\cite{qu2024explainable} builds a large dataset (ETTD) using GPT-4o, employing a two-stage analysis paradigm and fusion mask prompts. These works primarily focus on constructing domain-specific SFT datasets for suboptimal fine-tuning, while neglecting alignment with human preferences. Our method introduces a human-aligned preference dataset, Holmes-DPOSet, and employs Direct Preference Optimization~\cite{rafailov2024direct} to address the suboptimal fine-tuning issue.
\begin{table}[t]
\centering
\vspace{-0.1em}
\resizebox{0.49\textwidth}{!}{
\begin{tabular}{c|c|c|c|c}
\toprule[1.5pt]
 & \textbf{\#Generators} & \;\;\;\;\textbf{\#Image}\;\;\;\; & \textbf{Explanation} & \;\textbf{Pref. Data}\; \\ \hline
 \hline
CNNDetection~\cite{wang2019cnngenerated} & 11 & 720K & $\times$ & $\times$ \\ \hline
GenImage~\cite{zhu2023genimage} & 8 & 1M+ & $\times$ & $\times$ \\ \hline
DRCT~\cite{pmlr-v235-chen24ay} & 16 & 2M & $\times$ & $\times$ \\ \hline
WildFake~\cite{hong2024wildfake} & 21 & 3.5M+ & $\times$ & $\times$ \\ \hline
FakeBench~\cite{li2024fakebench} & 10 & 6K & $\checkmark$ & $\times$ \\ \hline
LOKI~\cite{ye2024loki} & 10 & 3K & $\checkmark$ & $\times$ \\ \hline
Holmes-Set & 18 & 65K+4K & $\checkmark$ & $\checkmark$ \\ \bottomrule[1.5pt]
\end{tabular}
}
\vspace{-0.5em}
\caption{Comparison between AI-Generated Image Detection Datasets. ``Pref. Data'' refers to human-aligned preference data.}
\vspace{-1.8em}
\label{tab:data_annotation1}
\end{table}

Additionally, some works have explored simple multimodal methods for detecting AI-generated images. Liu \textit{et al.}~\cite{liu2023forgery} enhanced the generalization of the detection method by considering the text encoding embeddings of CLIP and introducing frequency-related adapters into the image encoder. AntiFakePrompt~\cite{chang2023antifakeprompt}, Bi-Lora~\cite{keita2024bi}, and Jia \textit{et al.}~\cite{jia2024can} redefine the detection task as a visual question-answering task, combining vision-language models to improve performance on unseen data. Our work not only achieves state-of-the-art generalization detection accuracy but also provides further clues as to why an image is or is not AI-generated.

\section{Method}

\subsection{Data Pipeline}
\label{data_pipeline}

\begin{figure*}[h]
\centering
\vspace{-2mm}
\includegraphics[width=0.95\textwidth]{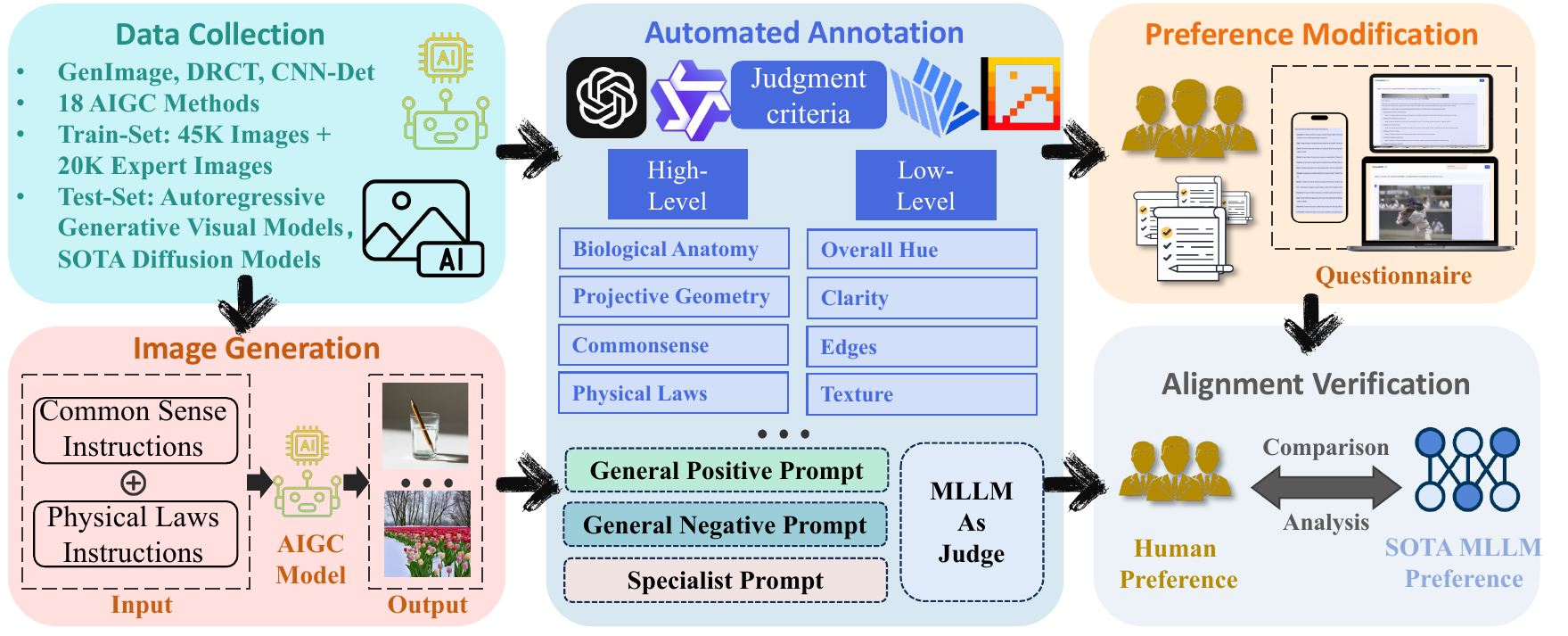}
\vspace{-4mm}
\caption{Details of the \textbf{Holmes-Set} Construction. The figure illustrates our data pipeline, consisting of four key components: Data Source (including Data Collection and Image Generation), Automated Annotation, Preference Modification (based on human expert feedback), and Comprehensive Evaluation (to assess model generalizability and interpretability). }
\label{fig:datapipeline}
\vspace{-4mm}

\end{figure*}

\noindent \textbf{Overview of Data Pipeline.} As shown in Fig.~\ref{fig:datapipeline}, our data pipeline consists of four components: Data Source, which includes Data Collection and Image Generation for gathering images to be annotated. These images are then sent to the Automated Annotation section. Subsequently, the annotations are refined in the Preference Modification stage based on expert feedback. Finally, we conduct a Comprehensive Evaluation to assess the model's generalizability and interpretability. In the following sections, we will introduce the detailed methodologies for each component in sequence.

\noindent \textbf{Data Source.} To ensure a diverse dataset with various types of forgeries and defects, as shown in Fig.~\ref{fig:datapipeline}, we initially selected 45K images from existing large-scale AI-generated image detection datasets such as CNNDetection~\cite{wang2019cnngenerated}, GenImage~\cite{zhu2023genimage}, and DRCT~\cite{pmlr-v235-chen24ay} for image description and forgery explanation. 
To reduce annotation costs, as in previous work~\cite{huang2024ffaa, xu2024fakeshield, sun2024forgerysleuth, qu2024explainable}, we consider using open-source multimodal large language models(MLLMs)  for annotation. A key challenge in constructing annotated explanation datasets by MLLMs for AI-generated images lies in two primary aspects. First, unlike deepfake detection and image forgery detection, AI-generated images often lack corresponding real images, complicating annotation efforts through MLLMs. Second, existing MLLMs exhibit limited capability in analyzing fine-grained forgeries within AI-generated images, as evidenced by recent studies~\cite{li2024fakebench}. This limitation likely stems from systematic gaps in domain-specific knowledge related to AI-generated image forensics within the training data of MLLMs. Such knowledge deficiencies may prevent MLLMs from reliably identifying common artifacts in AI-generated content, thereby introducing risks of incomplete or inaccurate annotations.

To mitigate the risk of the inaccurate and incomplete annotation results, inspired by \cite{kamali2024distinguish, kamali2025characterizing}, we employed expert-guided methods to filter out 20K images with common AI-generated defects, such as text~\cite{chen2024textdiffuser}, human bodies~\cite{fang2024humanrefiner}, human faces~\cite{huang2024ffaa}, projective geometry~\cite{Sarkar_2024_CVPR}, common sense~\cite{fu2024commonsense}, and physical laws~\cite{meng2024phybench}, from existing datasets. In simple terms, we use expert small models capable of identifying these defects to filter out the images, and then employ MLLMs to determine the presence of these defects. Additionally, due to the lack of attention to commonsense and physical law defects in existing datasets, we supplement the dataset with images generated from~\cite{fu2024commonsense, meng2024phybench}, as shown in the Image Generation part of Fig.~\ref{fig:datapipeline}. For these images containing common AI-generated defects, we can design targeted prompts to avoid potential annotation hallucinations, thereby reducing the need for extensive human labor in the annotation process. \textit{Further details of the image collection process are elaborated in the Appendix~\ref{supp:c}.}

\noindent \textbf{Automated Annotation.} We design the automated annotation system named Multi-Expert Jury, comprising four open-source multimodal large language models (MLLM-Experts): Qwen2VL-72B~\cite{Qwen2VL}, InternVL2-76B~\cite{chen2024far}, InternVL2.5-78B~\cite{chen2024expanding}, and Pixtral-124B~\cite{mistral_ai_pixtral_large}, to ensure high-quality data annotation. The system employs three tailored prompts:  1.~\textbf{General Positive Prompt} as shown in Fig.~\ref{fig:realprompt} and Fig.~\ref{fig:fakeprompt}: Annotates 45K randomly selected images using high-level (e.g., anatomy, physical laws) and low-level (e.g., texture, clarity) criteria (see Fig.~\ref{data_pipeline}).  2.~\textbf{General Negative Prompt} as shown in Fig.~\ref{fig:realprompt} and Fig.~\ref{fig:fakeprompt}: Generates adversarial annotations by asking questions contradicting image authenticity, forming natural positive-negative pairs with the General Positive Prompt annotations to construct the Direct Preference Optimization (DPO) dataset $\mathcal{D}_1$.  3.~\textbf{Specialist Prompt} as shown in Fig.~\ref{fig:face}, Fig.~\ref{fig:body}, Fig.~\ref{fig:text}, Fig.~\ref{fig:projective}, Fig.~\ref{fig:commonsense}: Guides MLLM-Experts to annotate 20K expert-filtered defective images, focusing on generation defects (e.g., commonsense). To ensure annotation quality, we adopt an MLLM-as-a-judge approach~\cite{chen2024mllm}, where MLLM-Experts cross-evaluate each annotation. Only annotations with top consensus scores are retained in the dataset.

\begin{figure*}[h!]
\vspace{-3mm}
\centering
\includegraphics[width=0.95\textwidth]{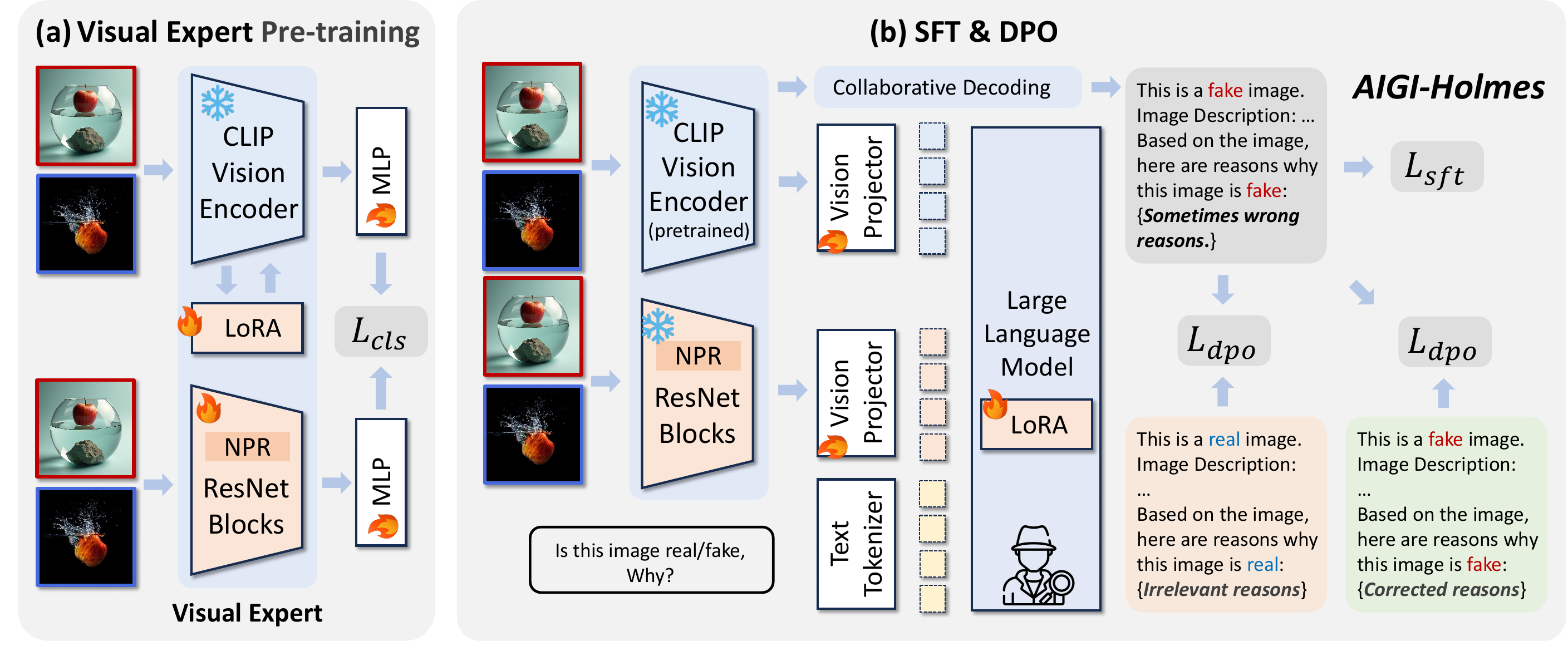}
\vspace{-3mm}
\caption{Overview of AIGI-Holmes. We enhance  LLaVA~\cite{liu2024visual} with NPR~\cite{tan2024rethinking} visual expert $\mathcal{R}$ and the Holmes Pipeline, featuring three training stages: Visual Expert Pre-training, SFT, DPO, and a collaborative decoding strategy during inference.}\label{fig:method}
\vspace{-5mm}
\end{figure*}

\noindent \textbf{Preference Modification.} The constructed SFT dataset can be directly used for coarse-grained alignment. However, there may be instances where models mechanically replicate explanation templates without genuinely understanding the underlying causes of artifacts or semantic errors. To further address this issue and narrow the gap between model perception and human reasoning, we introduce the Holmes-DPO dataset. We manually annotate an additional 2K data samples from the same sources as the training set. During this annotation process, we provide the outputs of the SFT model and ask humans to offer modification suggestions, such as supplementary correct information, removal of incorrect or irrelevant explanations, and other modification opinions. We then use Deepseek-V3~\cite{liu2024deepseek}, an advanced open-source large language model, to modify the original responses of the SFT model based on the suggestions of human experts, resulting in new, more human-aligned correct explanations. The human experts consist of three authors and two annotators who are strictly guided by us (we reference ~\cite{kamali2024distinguish, kamali2025characterizing} to create annotation documents for training and guiding the annotators). In addition, we use MLLMs and specialist prompts to modify responses to 2K filtered images with known specific defects. We add the pairs of samples before and after modification to the DPO dataset \(\mathcal{D}_2\).




\noindent \textbf{Comprehensive Evaluation.} We evaluate the model's capabilities from two perspectives: detection and explanation. \textbf{1. For detection,} we output the probabilities of the real/fake tokens to calculate Accuracy (Acc.) and Average Precision (A.P.). \textbf{2. For explanation,} we use 1K test samples containing Ground Truth, which have been reviewed by annotators for deficiencies in explanations from a professional perspective and corrected uniformly by the Deepseek-V3 model. We calculate metrics such as BLEU~\cite{papineni2002bleu}, CIDEr~\cite{vedantam2015cider}, METEOR~\cite{banerjee2005meteor}, and ROUGE~\cite{lin2004rouge} to measure the quality of the explanatory text output by the model. Additionally, we employ multimodal large model scoring and human preference evaluation methods for assessment. For MLLM scoring, we refer to~\cite{chen2024mllm}, using a prompt that considers relevance, accuracy, comprehensiveness, creativity, and granularity to compare and score the model's responses. For human evaluation scores, we sample 10 images for each type of forgery from $\mathcal{P}_3$, resulting in a total of 100 images. The comparative models generate explanations for these test images, and we use pairwise comparison, as referenced in~\cite{chiang2024chatbot}, to calculate the ELO ratings for each model's explanations.

\subsection{Overview of AIGI-Holmes}
As shown in Fig.~\ref{fig:method}, for the architecture, we augment the original multimodal method LLaVA~\cite{liu2024visual} with a low-level information NPR~\cite{tan2024rethinking} visual expert $\mathcal{R}$. For the training methodology, we introduce the Holmes Pipeline, which includes three key training stages: The visual Expert Pre-training Stage, the SFT Stage, and the DPO Stage. During the final inference, we employ a collaborative decoding strategy to merge the predictions from the visual expert and the language model.


\subsection{Architecture}
\label{arch}
Many existing methods use LLaVA~\cite{liu2024visual}, a Multimodal Large Language Model (MLLM) with strong multimodal understanding capabilities, as a baseline. However, this architecture faces several challenges: 1. Limited performance when dealing with classification problems~\cite{zhang2024visually}. 2. Inefficiency in handling low-level information~\cite{wu2023q}, which is crucial for AIGI detection. To address these challenges, we augment the CLIP visual encoder $\mathcal{F}$ used in LLaVA with a low-level information NPR~\cite{tan2024rethinking} visual expert $\mathcal{R}$. Given an image input \(X_{{img}}\), we extract visual features \(f_{{img}}\) and \(f_{{npr}}\) through $\mathcal{F}$ and $\mathcal{R}$, respectively, as shown in the following equations:
\vspace{-0.5em}
\begin{equation}
{
\footnotesize
\begin{aligned}
X_{{npr}} = \text{NPR}(X_{{img}});~
f_{{npr}} = \mathcal{R}(X_{{npr}});~
f_{{img}} = \mathcal{F}(X_{{img}}). 
\end{aligned}
}
\label{eq_1}
\vspace{-0.3em}
\end{equation}

\noindent Subsequently, the features are injected into the large language model through a projector and text embedding \(f_{{t}}\), producing the final output \(\mathbf{H}\). This process is represented by the following equations:
\vspace{-0.5em}
\begin{equation}
\mathbf{H} = \text{LLM}({\text{proj}}([f_{{img}}, f_{{npr}}]), f_{{t}}). 
\label{eq_2}
\end{equation}


%

\begin{figure*}[h]
\vspace{-0.3em}
\centering
\includegraphics[width=0.99\textwidth]{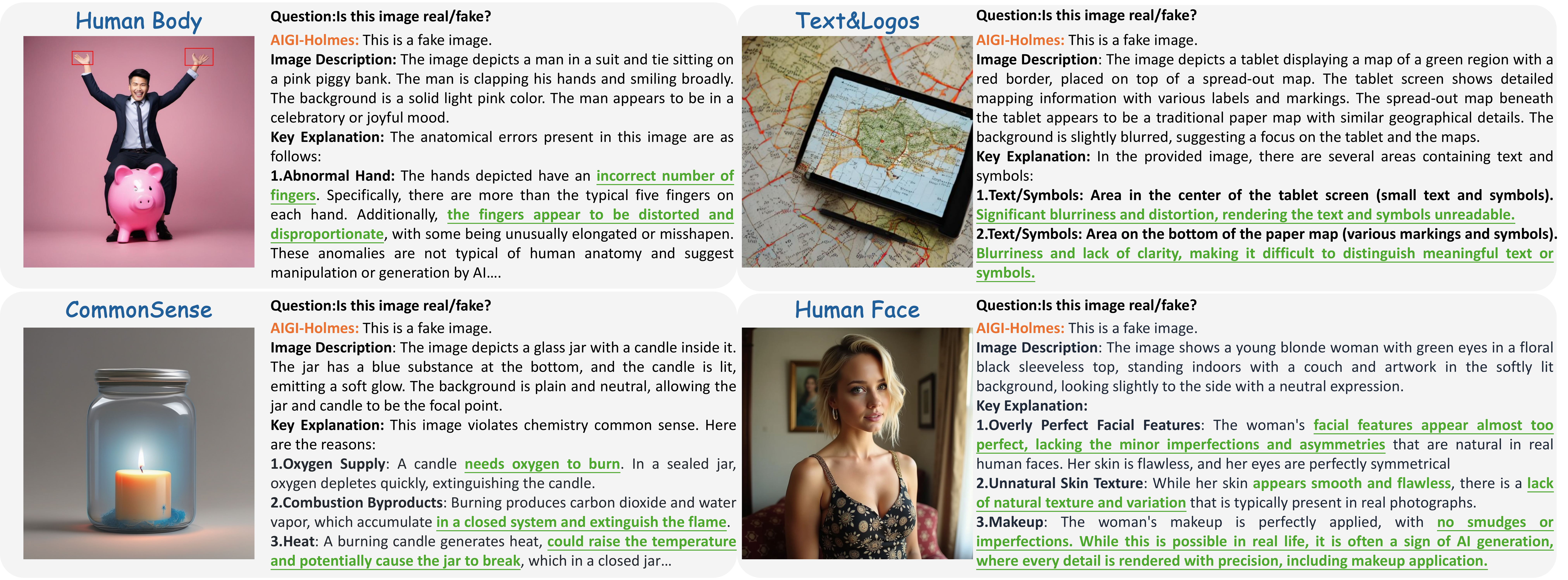}
\vspace{-3mm}
\caption{Qualitative results of AIGI-Holmes on AI-Generated images.}
\label{fig:example}
\vspace{-5mm}
\end{figure*}
\subsection{Holmes Pipeline}
\label{training}

\noindent \textbf{Visual Expert Pre-training Stage:} To transform the MLLM into an expert in the AIGI detection domain, we need to ensure that the visual expert provides a certain level of generalization capability and detection accuracy. Therefore, before the general LLava training paradigm, we introduce a pre-training stage for the visual expert. We employ LoRA~\cite{hu2021lora} to efficiently fine-tune CLIP-ViT-L/14, setting \(r=4\) and \(\alpha=8\). The CLS features \(f_{cls}\) extracted by CLIP are fed into an MLP to obtain classification results. Simultaneously, we independently adjust the NPR-based ResNet. Following~\cite{tan2024rethinking}, we only use the first two layers of ResNet to extract the features \(f_{npr}\). Similarly, an MLP is used to obtain classification results. For ResNet, we perform full parameter fine-tuning. We use binary cross-entropy loss \(l_{bce}\) to adjust the visual expert, as shown in the following equation:
\begin{equation}
\begin{aligned}
\mathbf{y}_{{clip}} &= \text{MLP}(f_{{cls}}),~~~~~~~~~ \mathbf{y}_{{npr}} = \text{MLP}(f_{{npr}}),  \\
l_{{clip}} &= l_{bce}(\mathbf{y}_{{clip}}, \mathbf{y}),~~~~~~~~~l_{{npr}} = l_{bce}(\mathbf{y}_{{npr}}, \mathbf{y}).
\end{aligned}
\label{eq_3}
\end{equation}


\noindent \textbf{SFT Stage}: After obtaining the pre-trained visual expert, we integrate it into the large language model. To guide the model to output explanations related to whether an image is AI-generated, we perform SFT on the Holmes-SFT dataset. In this stage, consistent with previous methods, we freeze the visual expert while keeping the linear projector and the LoRA components of the large language model trainable. We optimize these parameters using the autoregressive text loss \( l_{txt} \):
\begin{equation}
l_{txt} = l_{ce}(\mathbf{H}, \mathbf{H}_{txt}).
\label{eq_5}
\end{equation}
\vspace{-0.3em}

\noindent \textbf{DPO Stage}: To further enhance the ability of the SFT model to produce high-quality, human-aligned explanations, we perform human-aligned direct preference optimization on the previously constructed Holmes-DPOSet. Specifically, we sample preference pairs \(\{y_{w}\}, \{y_{l}\} \sim \mathcal{D}\) from the Holmes-DPOSet, where \(\mathcal{D} = \mathcal{D}_1 \cup \mathcal{D}_2\). We then optimize the model using the DPO loss as shown in Eq.~\ref{eq_dpo1}.
\begin{equation}
\small
\begin{aligned}
\mathcal{L}_{\text{DPO}}(\phi) &= -\mathbb{E}_{(x, y_{w}, y_{l}) \sim \mathcal{D}} \bigg[ \\
&\quad \log \sigma \bigg( \beta \Big( \log \frac{\pi_\phi(y_{w} | x)}{\pi_{\text{base}}(y_{w} | x)} - \log \frac{\pi_\phi(y_{l} | x)}{\pi_{\text{base}}(y_{l} | x)} \Big) \bigg) \bigg],
\end{aligned}
\label{eq_dpo1}
\end{equation}

\noindent where \(\sigma\) and \(\beta\) have the same meanings as in the first round. \(\pi_\phi(y | x)\) denotes the policy model to be optimized in the second round, with parameters \(\phi\). \(\pi_{\text{base}}(y | x)\) is the model after the first round of preference optimization. During the preference optimization stage, we keep the visual expert frozen, allow the projector to be trainable, and use LoRA to train the large language model part.
\begin{table*}[!h]
 \vspace{-2em}
    \centering
    \renewcommand\arraystretch{1.05}
    \setlength{\tabcolsep}{1.02mm}
      \resizebox{0.95\textwidth}{!}{
    \begin{tabular}{l  c c c c c c c c c c c c c c c c c c c c | c c}
    \toprule[1.5pt]
    \multirow{2}*{Method} & \multicolumn{2}{c}{Janus} & \multicolumn{2}{c}{Janus-Pro-1B} & \multicolumn{2}{c}{Janus-Pro-7B} & \multicolumn{2}{c}{Show-o} & \multicolumn{2}{c}{LlamaGen} & \multicolumn{2}{c}{Infinity} & \multicolumn{2}{c}{VAR} & \multicolumn{2}{c}{PixArt-XL} & \multicolumn{2}{c}{SD3.5-Large} & \multicolumn{2}{c|}{FLUX} & \multicolumn{2}{c}{Mean} \\
         \cline{2-23} 
~       & Acc. & A.P. & Acc. & A.P. & Acc. & A.P. & Acc. & A.P. & Acc. & A.P. & Acc. & A.P. & Acc. & A.P. & Acc. & A.P. & Acc. & A.P. & Acc. & A.P. & Acc. & A.P. \\ 
\bottomrule 
\hline
CNNSpot          & 70.0 & 86.0 & 70.9 & 85.8 & 85.0 & 93.6 & 72.2 & 86.0 & 61.9 & 71.4 & 86.8 & 94.6 & 59.9 & 75.0 & 78.2 & 90.1 & 63.8 & 81.1 & 79.9 & 92.0 & 72.9 & 85.6 \\
AntifakePrompt           & 72.2 & 87.4 & 84.3 & 94.0 & 84.8 & 93.1 & 86.2 & 95.5 & 96.2 & 99.4 & 83.6 & 94.1 & 90.7 & 95.6 & 81.7 & 92.8 & 92.8 & 97.8 & 66.1 & 80.8 & 83.9 & 93.1 \\
UnivFD          & 87.6 & 97.8 & 96.9 & 99.5 & 96.4 & 99.5 & 85.9 & 97.4 & 93.1 & 98.6 & 79.2 & 96.2 & 64.3 & 85.9 & 75.7 & 94.4 & 87.8 & 97.8 & 69.6 & 91.4 & 83.6 & 95.9 \\
NPR        & 51.2 & 55.9 & 69.5 & 75.1 & 73.9 & 77.9 & 93.7 & 99.6 & 93.5 & 99.4 & 93.8 & 99.9 & 85.9 & 91.2 & 93.4 & 99.1 & 91.6 & 97.7 & 93.6 & 99.5 & 84.0 & 89.5 \\
LaRE           & 70.8 & 99.3 & 74.7 & 97.5 & 95.6 & 99.7 & 80.0 & 99.0 & 91.6 & 99.6 & 77.9 & 99.6 & 98.8 & 100.0 & 82.2 & 99.7 & 94.1 & 99.5 & 84.3 & 99.0 & 85.0 & 99.3 \\
RINE & 89.9 & 98.3 & 98.7 & 99.9 & 97.2 & 99.6 & 98.8 & 99.9 & 99.1 & 100.0 & 99.2 & 99.9 & 85.0 & 97.9 & 98.9 & 99.8 & 97.8 & 99.7 & 97.1 & 99.7 & 96.2 & 99.5 \\
AIDE          & 91.2 & 99.1 & 98.9 & 99.9 & 97.8 & 99.8 & 98.0 & 99.8 & 99.4 & 100.0 & 98.7 & 99.9 & 93.6 & 99.3 & 98.6 & 99.9 & 99.4 & \textbf{100.0} & 94.4 & 99.5 & 97.0 & 99.7 \\
\bottomrule 
\hline
AIGI-Holmes           & \textbf{97.3} & \textbf{99.9} & \textbf{99.0} & \textbf{99.9} & \textbf{98.0} & \textbf{99.9} & \textbf{99.8} & \textbf{99.9} & \textbf{99.9} & \textbf{100.0} & \textbf{99.9} & \textbf{100.0} & \textbf{99.6} & \textbf{100.0} & \textbf{99.9} & \textbf{100.0} & \textbf{99.4} & {99.9} & \textbf{98.7} & \textbf{99.7} & \textbf{99.2} & \textbf{99.9} \\

\bottomrule[1.5pt]
    \end{tabular}
    }
    \vspace{-1.0em}
        \caption{Evaluation on the $\mathcal{P}_3$. All baseline results are trained on our training set to ensure a fair comparison.}
  \label{tab:p3}
        \vspace{-2em}
\end{table*}

\noindent \textbf{Inference Stage:} We propose \textbf{Collaborative Decoding}, which aims to utilize both our MLLM and its pre-trained expert to jointly decide the authenticity of an image during inference, thereby enhancing the generalization and detection accuracy of AIGI-Holmes. Specifically, we adjust the logit values of the tokens corresponding to ``real'' and ``fake'' in the model's output. We denote the logit value for ``real'' as \( \text{logit}(\mathbf{y}=0) \) and the logit value for ``fake'' as \( \text{logit}(\mathbf{y}={1}) \), for $k \in \{0,1\}$ as shown in Eq.~\ref{eq_4}:

{\footnotesize
\begin{equation}
\begin{aligned}
\text{logit}_{new}(\mathbf{y}=k) &= \alpha \cdot \text{logit}_{raw}(\mathbf{y}=k) \\
&\quad + \beta \cdot \text{logit}(\mathbf{y}_{{clip}}=k) + \gamma \cdot \text{logit}(\mathbf{y}_{{npr}}=k), \\
\end{aligned}
\label{eq_4}
\end{equation}
}

\noindent  where \(\alpha=1\), \(\beta=1\), and \(\gamma=0.2\) are the weights assigned to the three prediction results. Through collaborative decoding involving the pre-trained visual expert, we retain the predictions of the MLLM while preventing it from overfitting to existing forgery types, thereby improving the detection accuracy of the MLLM in unseen domains.

\section{Experiment}

\subsection{Experimental Setup}
\textbf{Datasets.} To comprehensively evaluate the generalization capabilities of existing methods, we conducted experiments under three settings: \textbf{Protocol-I}, \textbf{Protocol-II}, and \textbf{Protocol-III}. For \textbf{Protocol-I}, we trained on the 4-class (car, cat, chair, horse) subset of the CNNDetection dataset, which was widely used in earlier studies, and evaluated the detector on the general and comprehensive benchmark AIGCDetectBenchmark~\citep{zhong2023rich} ($\mathcal{P}_1$). For \textbf{Protocol-II}, we trained on the training set proposed by AntiFakePrompt~\cite{chang2023antifakeprompt} and tested on its proposed test set, which includes 18 types of forgeries. For \textbf{Protocol-III}, we trained on the dataset containing various Diffusion methods proposed in Sec.~\ref{data} and tested on images generated by the latest unseen autoregressive visual generation models Janus~\cite{wu2024janus}, Janus-Pro~\cite{chen2025janus}, VAR~\cite{VAR}, Infinity~\cite{Infinity}, Show-o~\cite{xie2024showo}, LlamaGen~\cite{sun2024autoregressive}, and the state-of-the-art diffusion models PixArt-XL~\cite{chen2024pixartsigma}, FLUX~\cite{flux}, SD3.5~\cite{esser2024scaling} ($\mathcal{P}_3$). Each of the above test sets contains 5K real images from COCO~\cite{lin2014microsoft} and 5K generated images corresponding to the generation methods.

\noindent \textbf{Implementation Details.} During the pre-training phase of the visual expert, we fine-tune CLIP-ViT/L-14 using LoRA (r = 4, $\alpha$ = 8) and fully fine-tune the first two layers of ResNet with NPR as input. The training is conducted for 5 epochs with a batch size of 32. During the SFT phase, we fine-tune LLaVA1.6-mistral-7B~\cite{liu2024llavanext} using LoRA (rank=128, $\alpha$=256) while fully training the domain label generator. This model is trained for 3 epochs with a learning rate of $5e-5$, a batch size of 16, and a gradient accumulation step of 1. During the DPO phase, we use LoRA (rank=48, $\alpha$=96) with a learning rate of $5e-7$, a batch size of 4, a gradient accumulation step of 2, and $\beta$=0.1, training for 2 epochs.

\noindent \textbf{Comparison Baselines.} We compare various baselines in the main text including CNNSpot~\cite{wang2020cnn}, NPR~\citep{tan2024rethinking}, AntiFakePrompt~\cite{chang2023antifakeprompt}, LaRE~\cite{luo2024lare}, RINE~\cite{koutlis2024leveraging}, AIDE~\cite{yan2024sanity}. \textit{Additional baseline methods will be introduced in the Appendix~\ref{supp:d1}.}


\subsection{Comparisons with SOTA Detection Methods}

\noindent \textbf{Protocol-III.} The quantitative results in Tab.~\ref{tab:p3} show the classification accuracy of various methods and generators within the range of $\mathcal{P}_3$. All methods were retrained on our proposed training set to ensure a fair comparison. The test images were generated by unseen state-of-the-art autoregressive visual models and diffusion models. On this challenging benchmark, AIGI-Holmes achieved SOTA results, with accuracy improvements of 15.2\%, 3.0\%, and 2.2\% over the previous best methods NPR, RINE, and AIDE, respectively. For the detection accuracy on the best autoregressive visual generation techniques and diffusion model representatives VAR and FLUX, our method surpassed the best methods by 6.0\% and 1.8\%, respectively. These three different training settings emphasize the excellent generalization ability of our proposed AIGI-Holmes.
\textit{The results of \textbf{Protocol-I} and \textbf{Protocol-II} can be found in the Appendix~\ref{supp:d3}.}





%
\begin{table*}[htbp]
\vspace{-0.3em}
\begin{minipage}{0.78\linewidth}
\centering
\renewcommand\arraystretch{1.}
\setlength{\tabcolsep}{1.25mm}
\resizebox{0.99\textwidth}{!}{%
\begin{tabular}{lcccccccccc}
\toprule[1.5pt]
\multirow{2}{*}{MLLM} & \multicolumn{5}{c}{Automatic Metrics} & \multicolumn{4}{c}{MLLM-as-Judge Evaluation} & \multirow{1}{*}{Human.} \\ \cmidrule(lr){2-6} \cmidrule(lr){7-10}
            & BLEU-1 & ROUGE-L & METEOR & CIDEr &  & Qwen2VL-72B & InternVL2-76B & InternVL2.5-78B & Pixtral-124B & ELO Ratings \\ \midrule
Qwen2VL-72B     & 0.314 & 0.227 & 0.292 & 0.003 &  & 3.874  & 3.612 & 4.002 & 3.163 & 8.432\\
InternVL2-76B & 0.362 & 0.224 & 0.289 & 0.006 &  & 4.042 & 3.807 & 4.006 & 3.463 & 10.111 \\
InternVL2.5-78B & 0.275 & 0.221 & 0.293 & 0.007 &  & 4.012 & 3.531 & 3.954 & 3.101 & 8.623 \\
Pixtral-124B & 0.428 & 0.270 & 0.302 & 0.010 &  & 3.967 & 3.990 & 4.140 & {4.213} & 10.472 \\ 
GPT-4o & 0.433 & 0.308 & 0.306 & 0.005 &  & 4.102 & 4.010 & 4.032 & {4.010} & 10.271 \\ \midrule
\rowcolor{gray!20} 
AIGI-Holmes(w/o DPO) & {0.445} & {0.315} & \textbf{0.317} & {0.023} &  & 4.119 & 3.918 & 4.150 & 4.130 & 10.670 \\
\rowcolor{gray!20} 
AIGI-Holmes(w/ DPO) & \textbf{0.622} & \textbf{0.375} & 0.311 & \textbf{0.107} &  & \textbf{4.196} & \textbf{4.011} & \textbf{4.189} & \textbf{4.227} & \textbf{11.420} \\
 \bottomrule[1.5pt]
\end{tabular}%
}
\vspace{-1em}
\caption{A comprehensive comparison of the explanations for AI-generated images between pre-trained SOTA MLLMs and AIGI-Holmes. The abbreviations ``w/o'' stands for ``without'', and ``w/'' stands for ``with''.} 
\label{table:exp}
\end{minipage}
\begin{minipage}{0.21\linewidth}
\centering
\renewcommand\arraystretch{1.3}
\setlength{\tabcolsep}{1.25mm}
\resizebox{0.9\textwidth}{!}{
\begin{tabular}{ccc|cc}
\toprule[1.5pt]
VEP-S & DPO & CD & $\mathcal{P}_1$ & $\mathcal{P}_3$ \\
\midrule[1pt]
            &            &                       & 83.3 & 90.1 \\
            &   \checkmark    &                       & 84.8 & 92.3 \\
\checkmark  &            &                       & 86.8 & 97.2 \\
\checkmark  & \checkmark &                       & 87.4 & 97.6 \\
\checkmark  &            & \checkmark            & 90.8 & 98.9 \\
\checkmark  & \checkmark & \checkmark            & \textbf{93.2} & \textbf{99.2} \\
\bottomrule[1.5pt]
\end{tabular}
}
\vspace{-0.5em}
\caption{Ablation Study of core model components.}
\label{tab_abl}
\end{minipage}
\end{table*}

\begin{table*}[t!]
\vspace{-1em}
\small
\begin{minipage}{0.51\linewidth}
\centering
\renewcommand\arraystretch{1.1}
\setlength{\tabcolsep}{1.25mm}
\resizebox{0.99\textwidth}{!}{
    \begin{tabular}{lccccc}
    \toprule[1.5pt]
    \multirow{2}[0]{*}{\textbf{Method}} & \multicolumn{2}{c}{\textbf{JPEG Compression}} & \multicolumn{2}{c}{\textbf{Gaussian Blur}} & \textbf{Resize} \\
          & \textbf{QF=75} & \textbf{QF=70} & \bm{$\sigma = 1.0$} & \bm{$\sigma = 2.0$} & \bm{$\times 0.5$} \\
          \midrule
    \textbf{CNNSpot} & 63.5  & 62.4  & 64.5  & 61.7  & 59.9 \\
    \textbf{NPR}   & 52.2  & 51.6  & 56.8  & 53.4  & 74.3 \\
    \textbf{UnivFD} &  84.7     &    84.0   &     81.0  &   74.9    &  86.3 \\
    \textbf{LaRE}  &   62.0    & 63.0      &     54.3  &    54.2   & 51.1 \\
    \textbf{AntifakePrompt} & 80.1  & 79.7  & 78.2  & 77.6  & 74.5 \\
    \textbf{AIDE}  & 92.8  & 92.3  & 91.9  & 90.7  & 89.2 \\
    \textbf{RINE}  & 92.4  & 91.1  & 94.2  & 92.8  & 92.3 \\
    \midrule
    \textbf{AIGI-Holmes} & \textbf{99.0}    & \textbf{98.7}  & \textbf{98.3}  & \textbf{97.9}  & \textbf{95.9} \\
    \bottomrule[1.5pt]
    \end{tabular}%
    }
\vspace{-0.6em}
\caption{{Robustness of Classification Accuracy on JPEG Compression, Gaussian Blur and Resize of AIGI-Holmes.} The classification accuracy (\%) averaged over 10 test sets in $\mathcal{P}_3$ with specific perturbation.}
\label{table:perturbation}
\end{minipage}
\hfill
\begin{minipage}{0.45\linewidth}
\vspace{8pt}
\centering
\vspace{-12pt}
\includegraphics[width=0.85\linewidth]{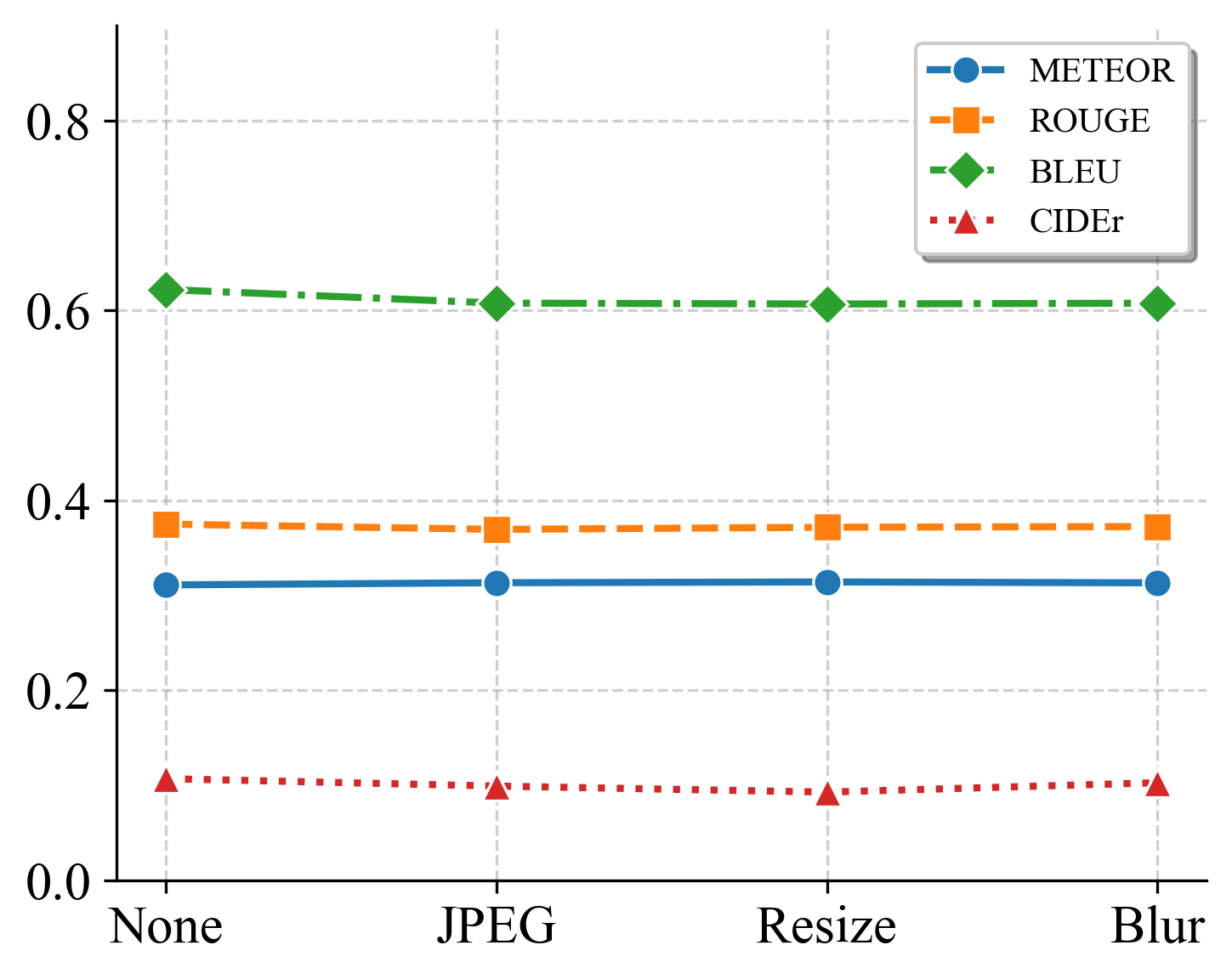}
\vspace{-8pt}
\captionof{figure}{Robustness of the explanation on JPEG Compression (QF=70), Gaussian Blur ($\sigma=2$), and Resize ($\times 0.5$) of AIGI-Holmes.}
\label{fig_rob_exp}
\end{minipage}
\vspace{-2.em}
\end{table*}

\subsection{Comparisons with SOTA MLLMs}
As demonstrated in Tab.~\ref{table:exp}, we conduct a quantitative comparison between the textual explanations generated by AIGI-Holmes and those produced by state-of-the-art (SOTA) multimodal large language models (MLLMs). To ensure a fair comparison, the explanations from the baseline models are obtained under the General Positive Prompt query. Our method achieves the highest metrics across nearly all evaluated aspects. For instance, compared to the state-of-the-art closed-source model GPT-4o, our model's output achieves a BLEU-1 score of 0.622, which is an improvement of 0.189 over GPT-4o's score of 0.433. Additionally, for human pairwise scoring, our model achieves an ELO Rating of 11.420, surpassing GPT-4o's score of 10.271 by 1.149 points. \textit{A selection of model outputs is illustrated in the Appendix~\ref{supp:e}.} This result indicates that existing MLLMs have the potential to provide reasonable explanations for AI-generated images. However, due to the lack of targeted downstream task training and the relatively small number of synthetic images included in the dataset, they are unable to perform more precise analyses in the task of explainable AIGI detection. Our method effectively fills this gap.

\subsection{Robustness Evaluation}

In real-world scenarios, AI-generated images often encounter unpredictable perturbations during dissemination, which can lead to the failure of existing AI detectors. Within the range of $\mathcal{P}_3$, we applied several common perturbations found in real-world scenarios: JPEG compression (QF=75, QF=70), Gaussian blur ($\sigma=1$, $\sigma=2$), and downsampling ($\times 0.5$). As shown in Tab.~\ref{table:perturbation}, the performance of all methods significantly declines under these distortions. However, AIGI-Holmes achieves higher detection accuracy compared to other baseline methods in these challenging scenarios. A possible reason is our use of the pre-trained CLIP method, which, as indicated in~\cite{cozzolino2024raising}, demonstrates good robustness when used as a backbone for AI image detection methods. Additionally, MLLM focuses more on high-level semantic features, reducing the model's reliance on low-level artifacts that are crucial for AI-generated image detection in other methods. These artifacts are often susceptible to unpredictable perturbations in real-world scenarios. Additionally, as shown in Fig.~\ref{fig_rob_exp}, under these degradation conditions, the evaluation metrics for model explanations such as BLEU-1, ROUGE-L, METEOR, and CIDEr did not exhibit significant declines. This indicates that the explanations generated by the model remain focused on high-level semantic information related to the image content and are not overly affected by these degradation conditions.

\subsection{Ablation Study}
We conduct ablation experiments on the main innovative methods of AIGI-Holmes: Visual Expert Pre-training Stage (VEP-S), DPO, and Collaborative Decoding (CD), as shown in Tab.~\ref{tab_abl}. The results demonstrate a significant improvement in accuracy when using the Visual Expert Pre-training Stage compared to the original llava paradigm, with detection accuracy increasing by 3.5\% and 7.1\% on $\mathcal{P}_1$ and $\mathcal{P}_3$, respectively. Adding DPO and Collaborative Decoding on this basis further enhances accuracy. Specifically, DPO helps improve the model's accuracy by 0.6\% and 0.4\% on $\mathcal{P}_1$ and $\mathcal{P}_3$, respectively, while Collaborative Decoding in conjunction with the Visual Expert Pre-training Stage boosts accuracy by 4.0\% and 1.7\%. The combined use of these approaches results in improvements over previous combinations, achieving approximately a 10\% increase in accuracy compared to the baseline method. These experiments demonstrate the effectiveness of our design. After employing the DPO stage, a comprehensive improvement in the quantitative results of model output explanations is observed, particularly with an increase of 0.75 points in human-scored ELO Ratings compared to the SFT model. This indicates that DPO is an effective post-training method for enhancing the quality of model output explanations to better align with human preferences. \textit{Results of other ablation experiments can be found in the Appendix~\ref{supp:d4}}.



\subsection{Qualitative Results}
Fig.~\ref{fig:example} showcases the detection and explanation results generated by AIGI-Holmes on various AI-generated images. It can be observed that our method produces precise explanations for defects across different AI-Generation modes. \textit{More visual results can be found in the Appendix~\ref{supp:e}.}


\section{Conclusion}
In this work, we introduce Holmes-Set, the first explanation-rich dataset with human-verifiable semantic annotations and contrastive preference pairs, addressing the critical data scarcity in AI-generated image detection. Besides, the proposed Holmes Pipeline systematically integrates visual expert pre-training, explanation-aware fine-tuning, and human-aligned preference optimization, synergizing model perception with human reasoning. Extensive experiments demonstrate our AIGI-Holmes' state-of-the-art detection accuracy and human-aligned interpretability. These contributions advance explainable and generalizable AIGI detection for rapidly evolving AIGC scenarios. We hope our dataset and methodology will inspire future research toward building more trustworthy AI-generated image detection systems.



\section*{Acknowledgements}
This work was supported by the National Science Fund for Distinguished Young Scholars (No.62025603), the National Natural Science Foundation of China (No. U21B2037, No. U22B2051, No. U23A20383, No. U21A20472, No. 62176222, No. 62176223, No. 62176226, No. 62072386, No. 62072387, No. 62072389, No. 62002305 and No. 62272401), and the Natural Science Foundation of Fujian Province of China (No. 2021J06003, No.2022J06001).

{
    \small
    \bibliographystyle{ieeenat_fullname}
    \bibliography{main}
}
\appendix
\clearpage
\setcounter{page}{1}
\maketitlesupplementary

\begin{algorithm}
\small
\caption{Holmes-Set Construction}
\begin{algorithmic}[1]
\Require Source datasets $\mathcal{M}$, Expert datasets $\mathcal{E}$, MLLM jury $\mathcal{J}$, SFT budget $K$, DPO rounds $R_1, R_2$
\Statex
\State // Stage 1: Data Collection
\State $\mathcal{D}_{\text{base}} \leftarrow \text{Select}(\mathcal{M}, 45\text{K})$ \Comment{From CNNDetection, GenImage, etc.}
\State $\mathcal{D}_{\text{expert}} \leftarrow \text{Filter}(\mathcal{E}, 20\text{K})$ \Comment{Expert-guided flaws}
\State $\mathcal{D}_{\text{gen}} \leftarrow \text{Generate}(\cite{fu2024commonsense, meng2024phybench})$  \Comment{Common sense/physics flaws}
\State $\mathcal{D}_{\text{all}} \leftarrow \mathcal{D}_{\text{base}} \cup \mathcal{D}_{\text{expert}} \cup \mathcal{D}_{\text{gen}}$
\Statex
\State // Stage 2: Automated Annotation
\State $\mathcal{D}_{\text{SFT}}, \mathcal{D}_1 \leftarrow \emptyset$
\For{each image $I \in \mathcal{D}_{\text{all}}$}
    \If{$I \in \mathcal{D}_{\text{expert}} \cup \mathcal{D}_{\text{gen}}$}
        \State $x_p \leftarrow \text{SpecialistPrompt}(I)$ \Comment{Focus on known flaws}
    \Else
        \State $x_p \leftarrow \text{GeneralPositivePrompt}()$ \Comment{Judge criteria}
        \State $x_n \leftarrow \text{GeneralPositivePrompt}()$ 
    \EndIf
    \State $A \leftarrow \mathcal{J}(I, x_p)$ \Comment{Jury system generates annotation}
    \State $A_{-} \leftarrow \mathcal{J}(I, x_n)$ \Comment{Jury system generates annotation}

    \Statex
    \State // Self-verification for SFTSet
    \State $\text{score} \leftarrow \frac{1}{|\mathcal{J}|} \sum_{J \in \mathcal{J}} \text{Score}(A, J(I))$
    \If{$\text{score} \geq \delta_{\text{SFT}}$}
        \State $\mathcal{D}_{\text{SFT}} \leftarrow \mathcal{D}_{\text{SFT}} \cup \{(I, A)\}$
    \EndIf
    \State $\mathcal{D}_1 \leftarrow \mathcal{D}_1 \cup \{(I, A, A_{-})\}$
\EndFor
\Statex
\State // Stage 3: Human Preference Refinement
\State $\mathcal{D}_2 \leftarrow \emptyset$
\State Sample $\mathcal{D}_{\text{human}} \subset \mathcal{M} \cup \mathcal{E}$ (2K images)
\State Sample $\mathcal{D}_{\text{mllm}} \subset \mathcal{M} \cup \mathcal{E}$ (2K images)
\For{each $(I, A) \in \mathcal{D}_{\text{human}}$}
    \State $A_1' \leftarrow \text{HumanRevise}(A)$ \Comment{Human-Expert modifications}
    \State $A' \leftarrow \text{DeepseekV3}(A, A_1')$ \Comment{LLM-assisted refinement}
    \State $\mathcal{D}_2 \leftarrow \mathcal{D}_2 \cup \{(I, A')\}$
\EndFor
\For{each $(I, A) \in \mathcal{D}_{\text{mllm}}$}
    \State $A_2' \leftarrow \text{MLLMRevise}(A)$ \Comment{MLLM-Expert modifications}
    \State $A'' \leftarrow \text{DeepseekV3}(A, A_2')$ \Comment{LLM-assisted refinement}
    \State $\mathcal{D}_2 \leftarrow \mathcal{D}_2 \cup \{(I, A'')\}$
\EndFor

\Statex
\State // Stage 4: Comprehensive Evaluation
\State $\mathcal{D}_{\text{test}} \leftarrow \text{HumanRevise}(1\text{K})$
\State Compute $\{\text{BLEU, CIDEr, ROUGE-L, METER}\}$ on $\mathcal{D}_{\text{test}}$
\State $\text{MLLM-Score} \leftarrow \frac{1}{|\mathcal{J}|} \sum_{J \in \mathcal{J}} \text{BatchScore}(\mathcal{D}_{\text{test}}, J)$
\State $\text{Human-Score} \leftarrow \text{ExpertPreferenceAssessment}(\mathcal{D}_{\text{test}})$
\Statex
\State \Return $\mathcal{D}_{\text{SFT}}, \mathcal{D}_1, \mathcal{D}_2$
\end{algorithmic}
\label{data_pipeline}
\end{algorithm}  

\section{Prompts}
\label{supp:a}
\begin{figure*}[h]
\centering
\includegraphics[width=0.9\textwidth]{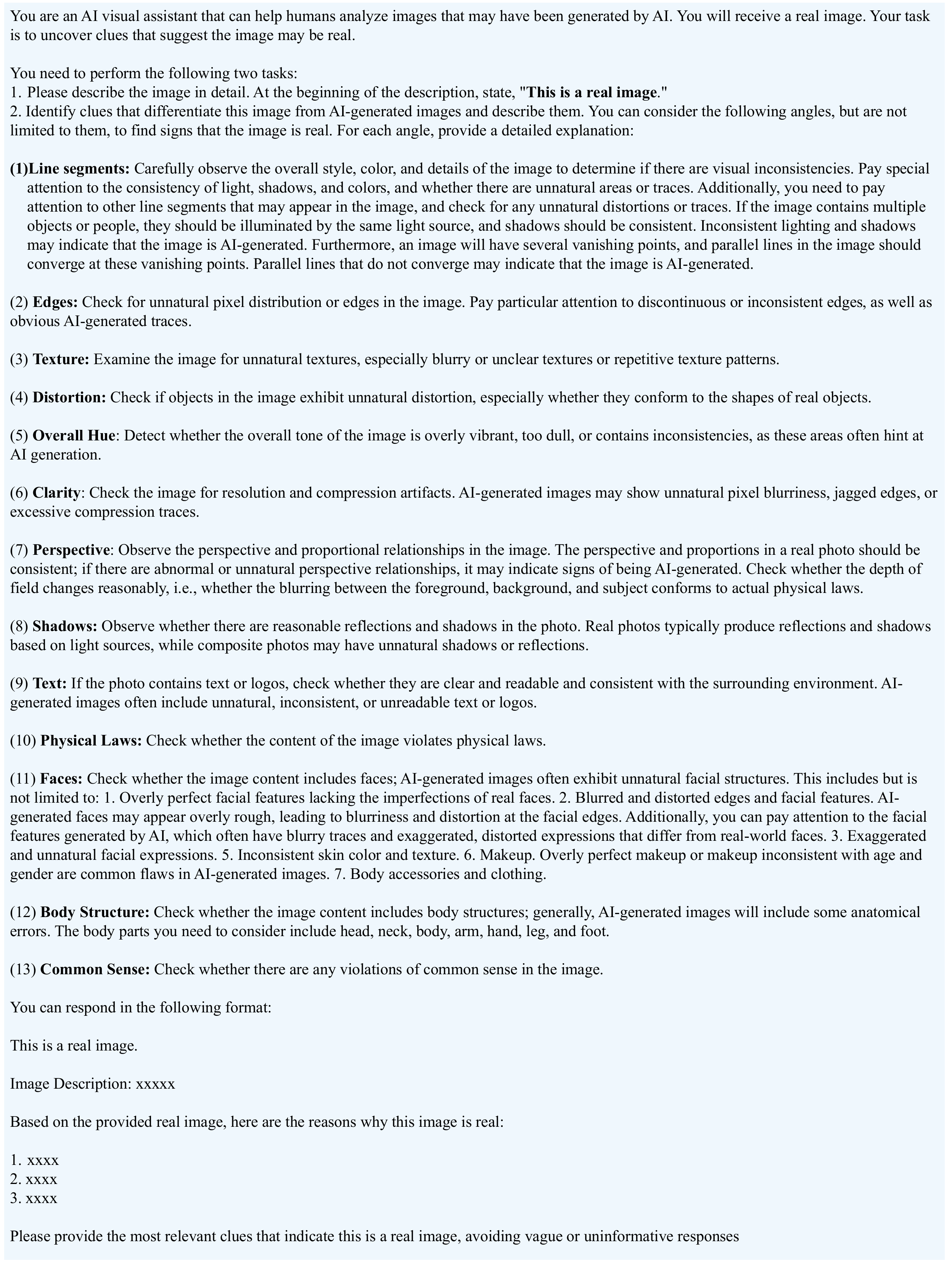}
\vspace{-3mm}
\caption{
General Positive Prompt for annotating real images and General Negative Prompt for annotating AI-generated images.}
\label{fig:realprompt}
\vspace{-8mm}
\end{figure*}

\begin{figure*}[h]
\centering
\includegraphics[width=0.9\textwidth]{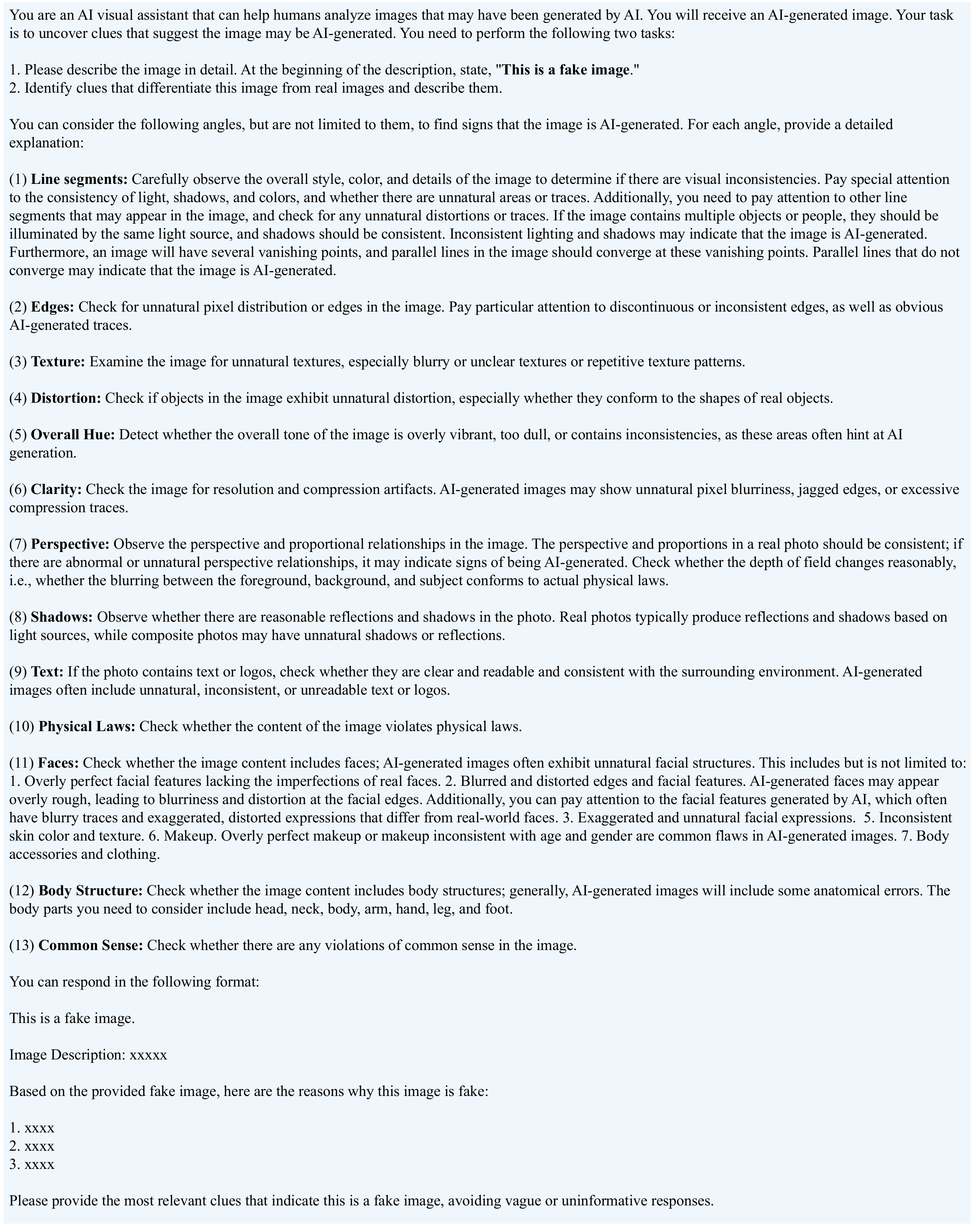}
\vspace{-3mm}
\caption{General Positive Prompt for annotating AI-generated images and General Negative Prompt for annotating real images.}
\label{fig:fakeprompt}
\vspace{-8mm}
\end{figure*}

\begin{figure*}[h]
\centering
\includegraphics[width=0.83\textwidth]{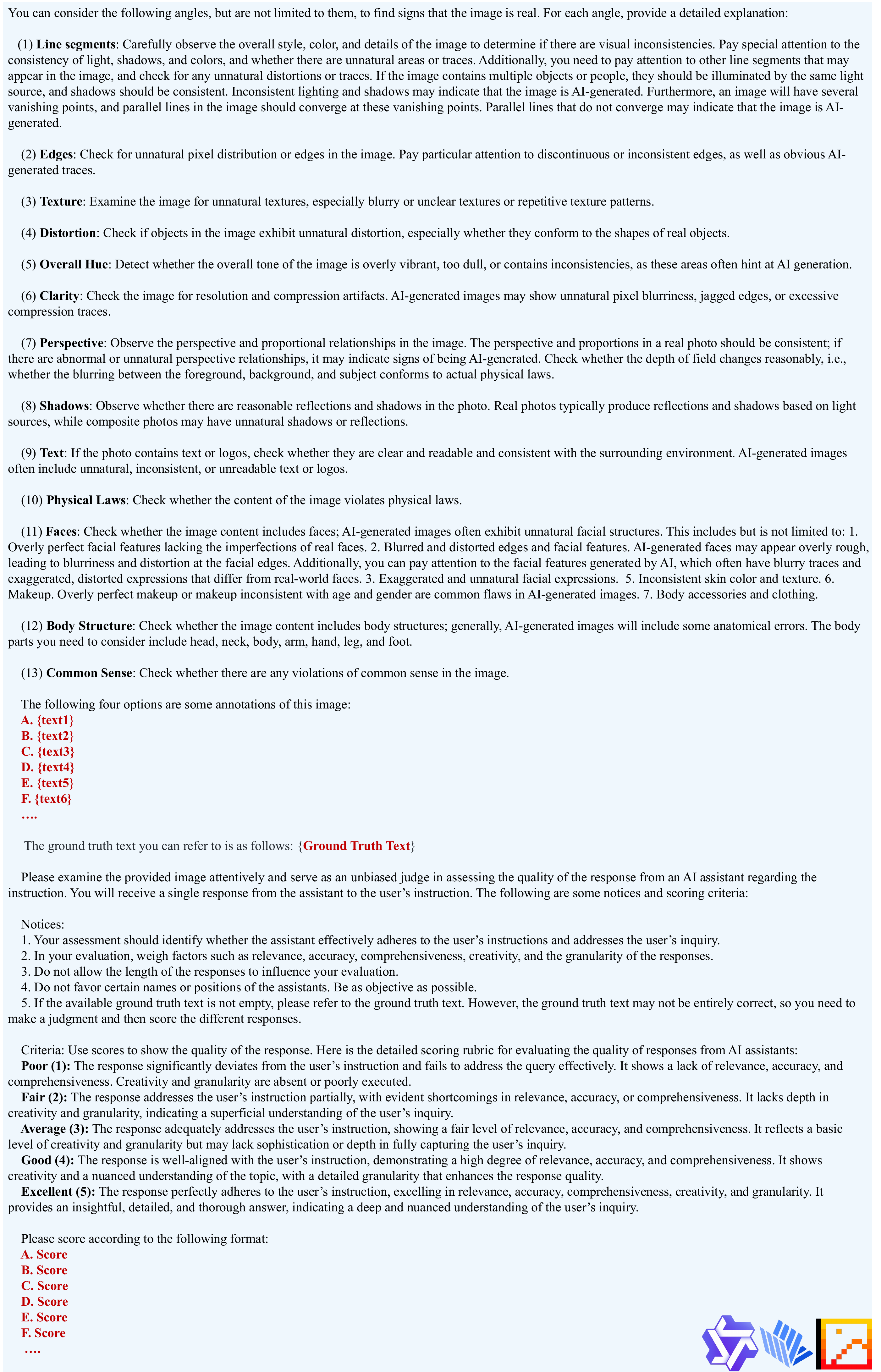}
\vspace{-3mm}
\caption{Prompt for cross-model validation and evaluation using state-of-the-art multimodal large language models.}
\label{fig:evaluate}
\vspace{-8mm}
\end{figure*}

We present the prompts used in our paper as follows: the prompt for querying real images is shown in Fig.~\ref{fig:realprompt}, and the prompt for querying AI-generated images is shown in Fig.~\ref{fig:fakeprompt}. The prompts for validation and evaluation in the Multi-Expert Jury are displayed in Fig.~\ref{fig:evaluate}. Prompts for filtering images with specific defects and their corresponding queries are shown in Fig.~\ref{fig:face}, Fig.~\ref{fig:body}, Fig.~\ref{fig:text}, Fig.~\ref{fig:projective}, Fig.~\ref{fig:commonsense}. The prompt for modifying DeepseekV3~\cite{liu2024deepseek} based on feedback is shown in Fig.~\ref{fig:dpsk_prompt}.


\section{Additional Related Work}
\label{supp:b}
In the Deepfake Detection, in addition to existing black-box binary classification methods~\cite{sun2022dual, sun2021domain, sun2022information, chen2024diffusionface, sun2025continual, chen2024diffusionfake}, a new wave of explainable detection approaches based on multimodal large models has emerged~\cite{zhang2024common, sun2025towards, guo2025rethinking, chen2024textitx2dfdframeworkexplainableextendable, lian2024large, yu2025unlocking, kundu2025truthlens, peng2025mllm}.  $X^2$DFD~\cite{chen2024textitx2dfdframeworkexplainableextendable} introduced expert-level features for better explanatory outputs, while ForgeryTalker~\cite{lian2024large} developed a framework for DeepFake facial image localization and explanation using a multimodal forgery tracking dataset (MMTT) and a forgery prompt network (FPN). In Image Forgery Detection, ForgeryGPT~\cite{li2024forgerygpt} integrated a Mask-Aware Forgery Extractor, improving detection and localization tasks (IFDL) with interpretable reasoning. SIDA~\cite{huang2024sida} proposed a framework for detecting deepfakes in social media images, which was trained on a dataset of 300,000 images (SID-Set), including 3,000 images with annotations. In AI-Generated Image Evaluation, HEIE~\cite{yang2024heie} introduced a CoT-driven evaluator and an irrationality mapper, combining heatmaps, scores, and explanations for fine-grained defect localization, supported by the ExplAIGI-Eval dataset. For AI-generated video detection, MM-Det~\cite{on-learning-multi-modal-forgery-representation-for-diffusion-generated-video-detection} extracted multimodal features from videos using LLaVA~\cite{liu2024visual} and designed a sophisticated fusion strategy. Notably, two contemporaneous works on explainable AI-generated image detection, LEGION~\cite{kang2025legion} and FakeVLM~\cite{wen2025spot}, have attracted our attention. We briefly compare these works with ours as follows: In terms of data construction, FakeVLM utilizes MLLM-generated explanations, while LEGION relies solely on human annotation. Our approach leverages MLLM-generated annotations with expert-guided filtering and human corrections, striking a balance between annotation cost and data quality. Regarding training, unlike FakeVLM and LEGION, which only employ supervised fine-tuning (SFT), we introduce a systematic three-stage training pipeline—visual-expert pre-training, SFT, and direct preference optimization (DPO)—to provide a more comprehensive exploration of adapting MLLMs for explainable AI-generated image detection.

\section{Holmes-Set Analysis}
\label{supp:c}
In the main text and Fig.~\ref{fig:datapipeline}, we introduced the specific construction process of the Holmes-Set. Furthermore, we present the pseudocode of the Holmes-Set construction process in Algorithm~\ref{data_pipeline} for reference. In the following sections, we will analyze the dataset and visualize some examples.

\begin{figure*}[h]
\centering
\includegraphics[width=0.72\textwidth]{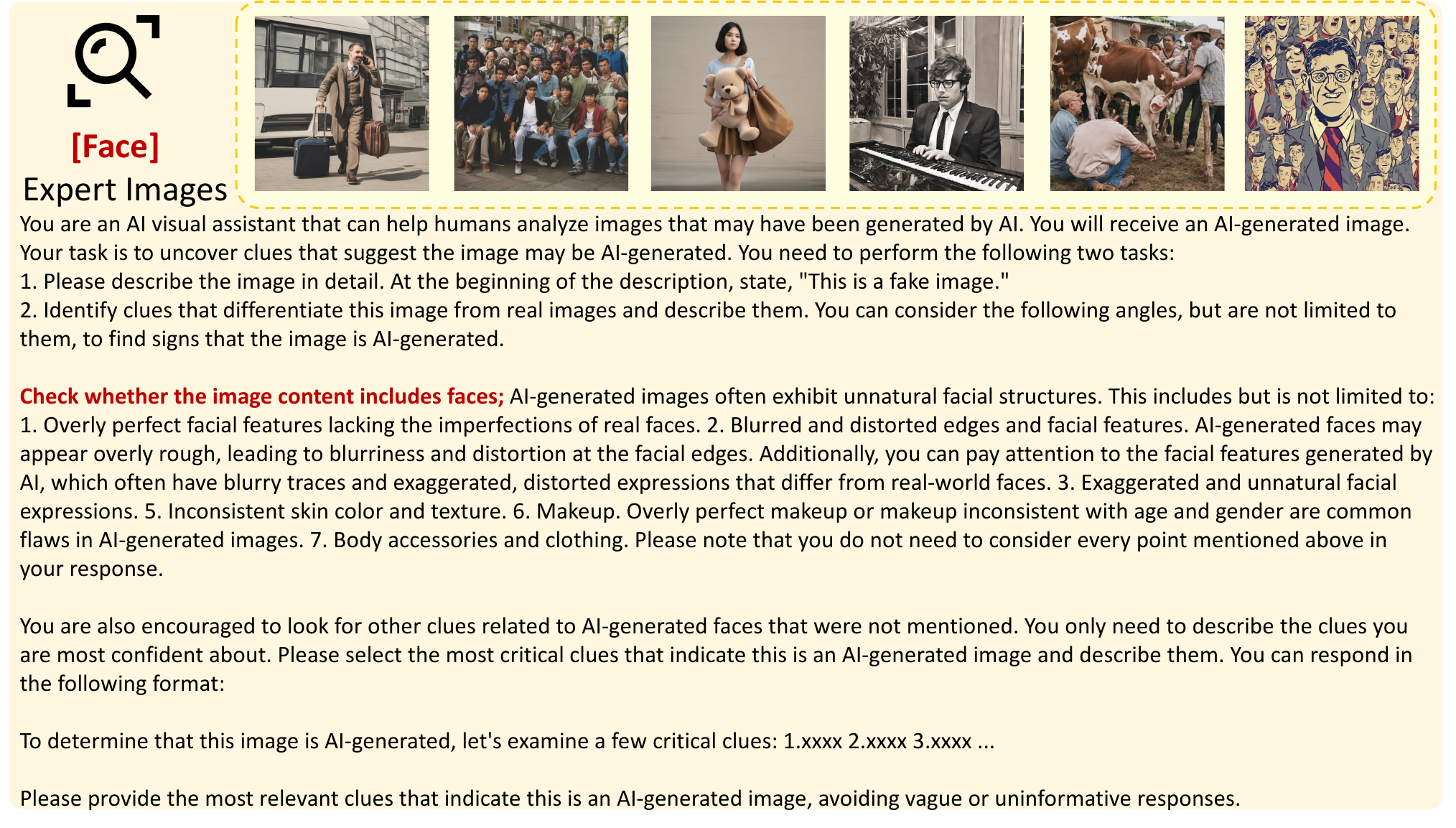}
\vspace{-3mm}
\caption{Specialist Prompt for AI-generated images containing face defects.}
\label{fig:face}
\vspace{-8mm}
\end{figure*}

\begin{figure*}[h]
\centering
\includegraphics[width=0.72\textwidth]{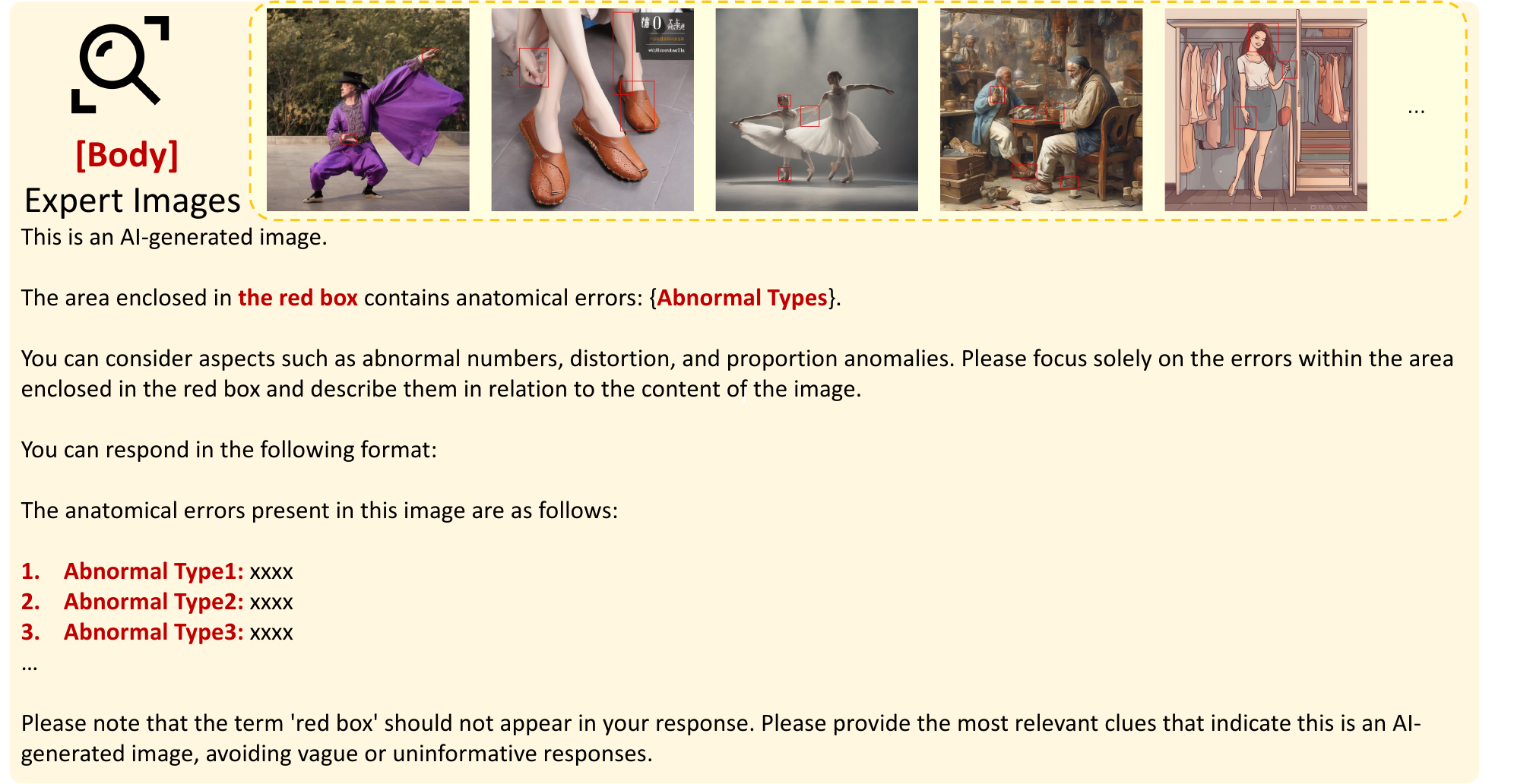}
\vspace{-3mm}
\caption{Specialist Prompt for AI-generated images containing body defects.}
\label{fig:body}
\vspace{-8mm}
\end{figure*}

\begin{figure*}[h]
\centering
\includegraphics[width=0.75\textwidth]{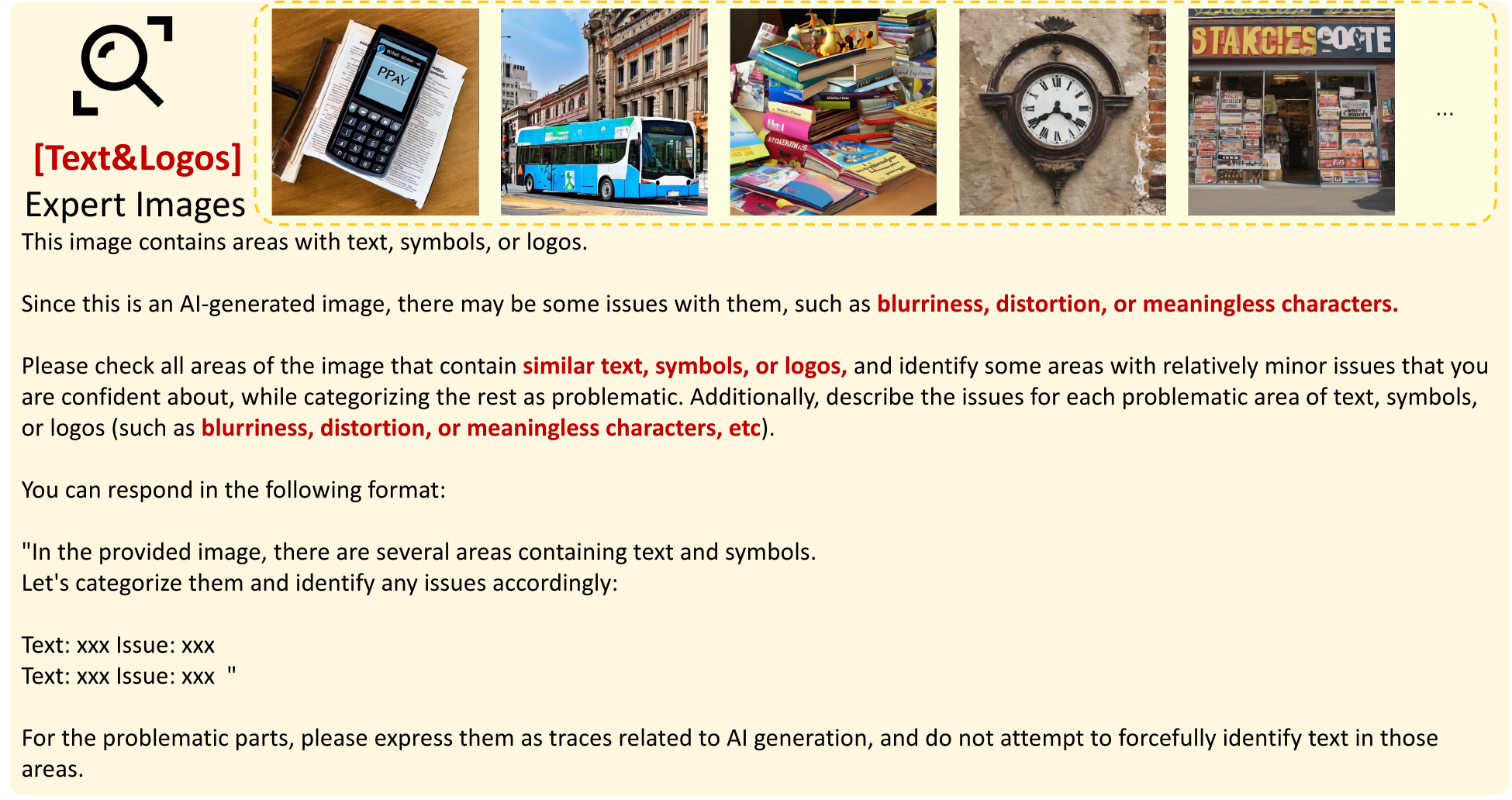}
\vspace{-3mm}
\caption{Specialist Prompt for AI-generated images containing defects in text\&logos.}
\label{fig:text}
\vspace{-8mm}
\end{figure*}

\begin{figure*}[h]
\centering
\includegraphics[width=0.75\textwidth]{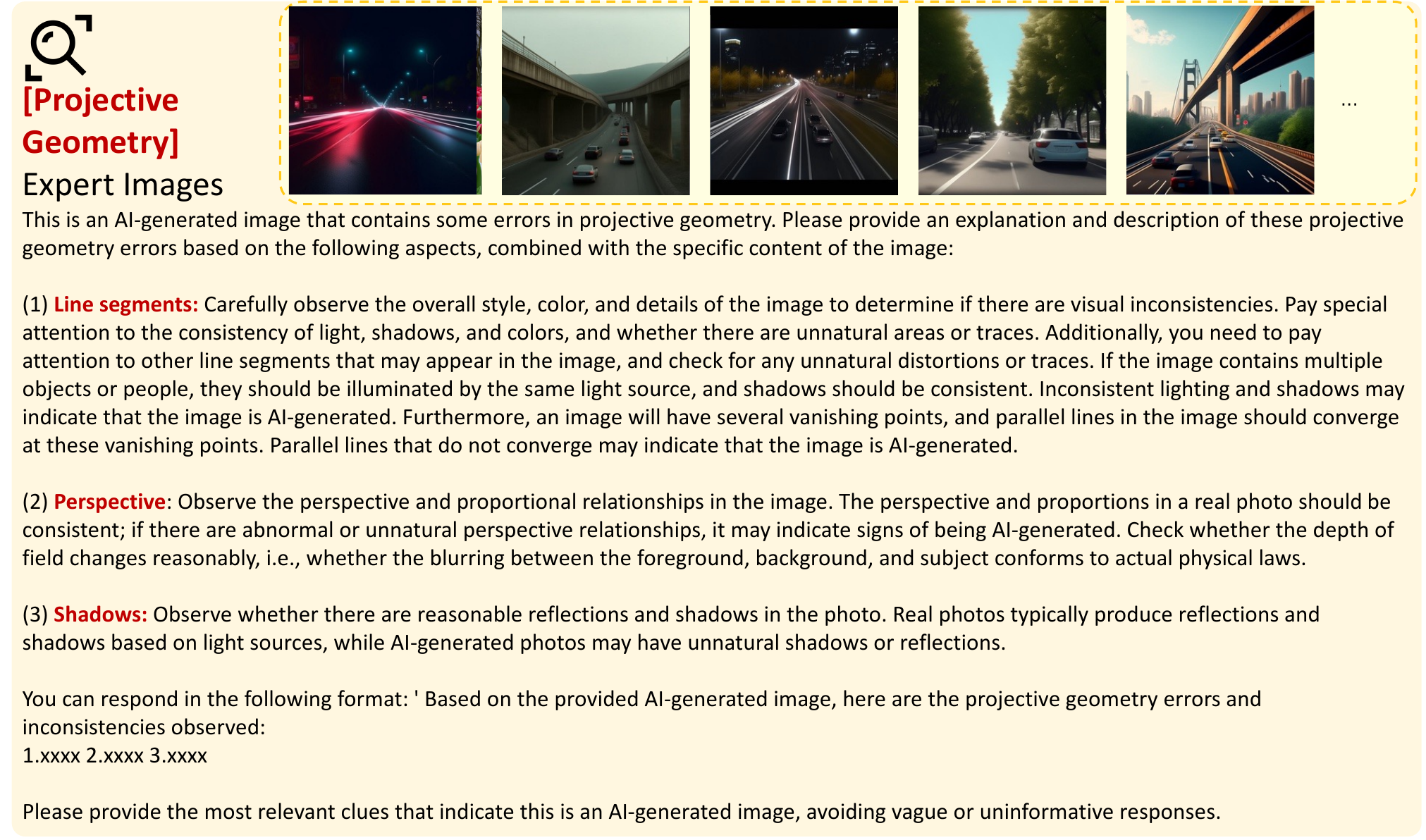}
\vspace{-3mm}
\caption{Specialist Prompt for AI-generated images containing defects in projective geometry.}
\label{fig:projective}
\vspace{-8mm}
\end{figure*}

\begin{figure*}[h]
\centering
\includegraphics[width=0.75\textwidth]{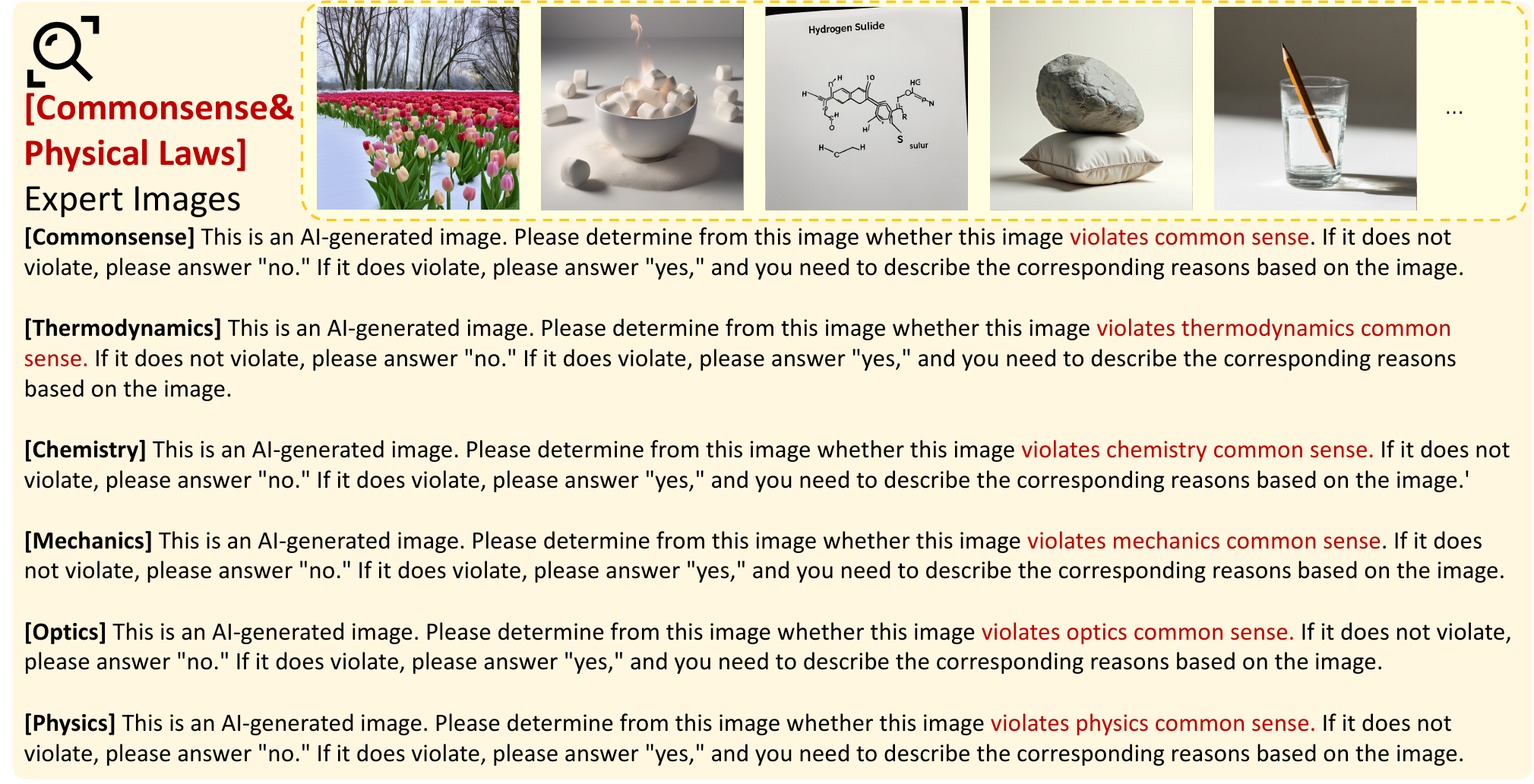}
\vspace{-3mm}
\caption{Specialist prompt for AI-generated images containing defects in commonsense and physical laws.}
\label{fig:commonsense}
\vspace{-8mm}
\end{figure*}

\noindent \textbf{Expert-guided Filter Method.} In the main text, we describe the process of filtering generated images that contain common AI-generated flaws. Below are the expert models we used:
\begin{itemize}
\item \textbf{Face}: We used a face detection method from OpenCV2 to filter out images containing faces from AI-generated images. MLLMs were also used for annotation and verification, as shown in Fig.~\ref{fig:face}.
\item \textbf{Human Body}: We obtained images with various body anomalies from AbHuman~\cite{fang2024humanrefiner} and generated red bounding boxes based on the provided bounding boxes to serve as visual prompts for MLLMs, as depicted in Fig.~\ref{fig:body}.
\item \textbf{Text \& Logos}: We utilized a model from PaddleOCR\footnotemark[1] to filter out images containing text. Additionally, we employed MLLMs for annotation and verification, as illustrated in Fig.~\ref{fig:text}.
\item \textbf{Projective Geometry}: We used the model from~\cite{Sarkar_2024_CVPR} to filter out images that might contain projective geometry errors. MLLMs were employed for annotation and verification, as illustrated in Fig.~\ref{fig:projective}.
\item \textbf{Common Sense \& Physical Laws}: We generated a batch of images using prompts that induce hallucinations in T2I generation models, as mentioned in commonsense~\cite{fu2024commonsense} and physical laws~\cite{meng2024phybench}. MLLMs were used for annotation and verification, as shown in Fig.~\ref{fig:commonsense}.
\end{itemize}
For each type of flaw mentioned above, we ultimately obtained 2,000 annotated image-text pairs, and selected 2,000 images from the COCO dataset as the corresponding real image dataset.

\begin{figure*}[h]
\centering
\includegraphics[width=0.99\textwidth]{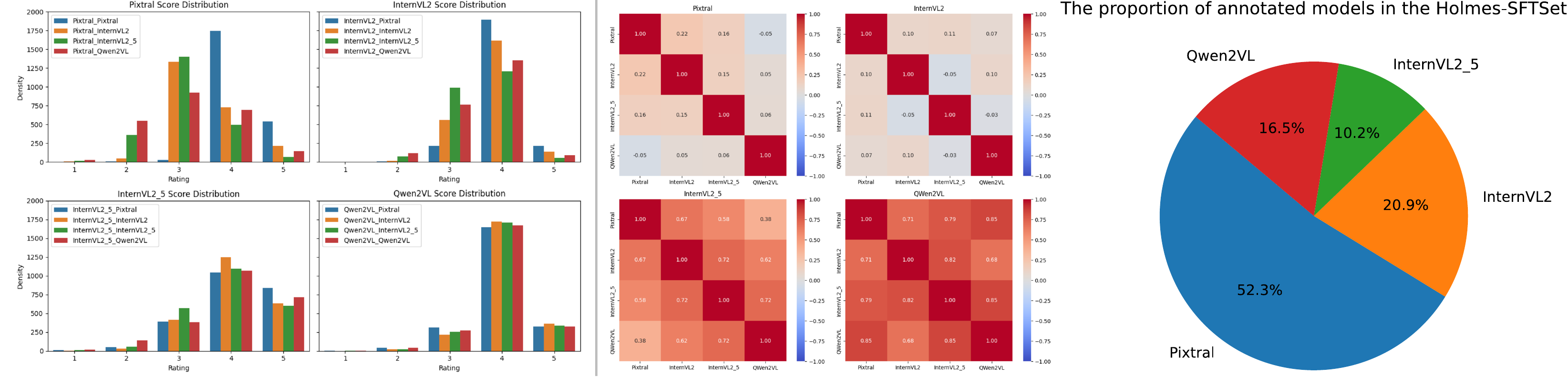}
\vspace{-3mm}
\caption{The score distribution of Multi-Expert Jury ratings, the correlation heatmap of the ratings, and the proportion of annotations by each MLLM Expert in Holmes-SFTSet.}
\label{fig:stat}
\vspace{-8mm}
\end{figure*}

\begin{figure*}[h]
\centering
\includegraphics[width=0.99\textwidth]{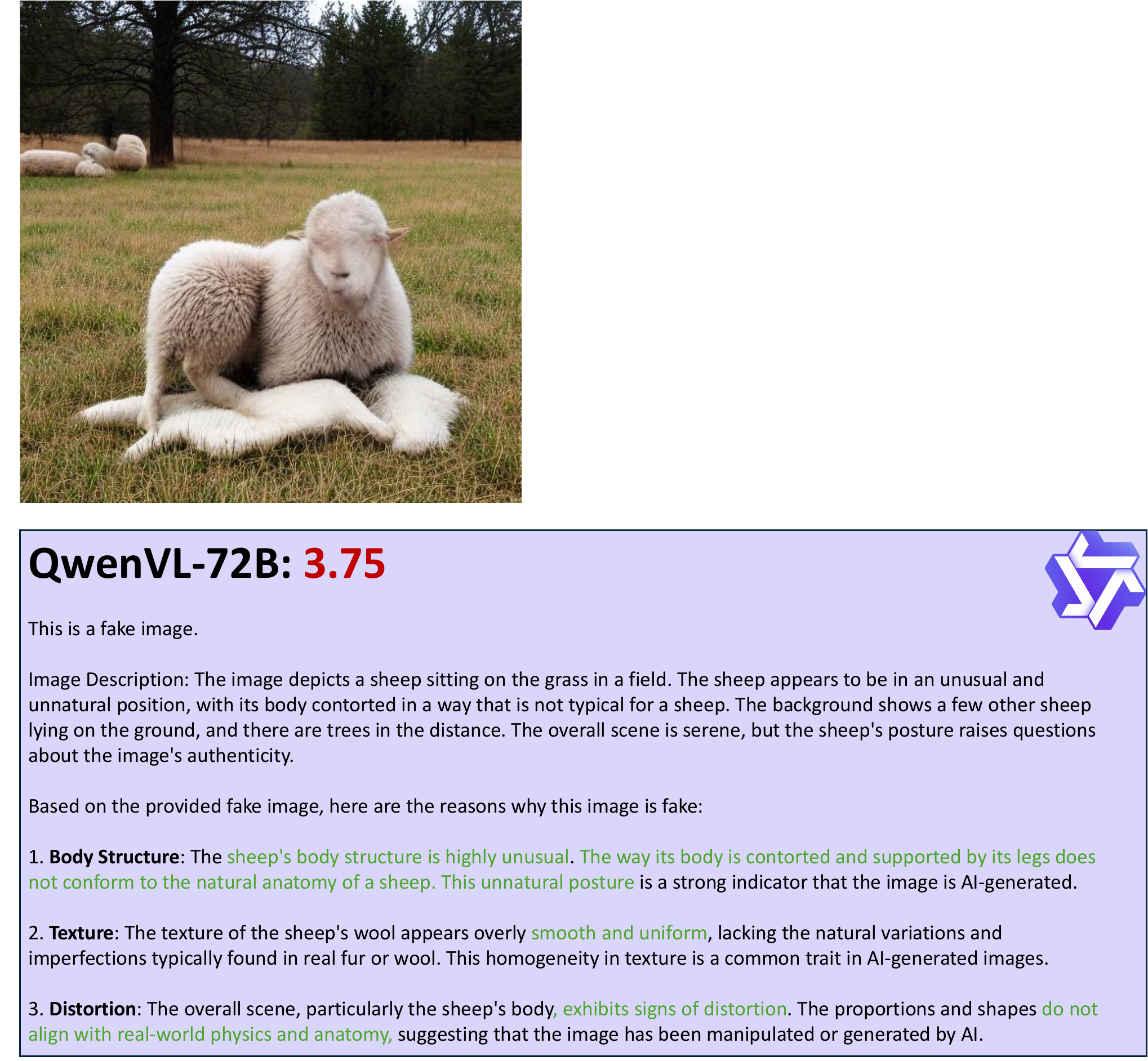}
\vspace{-3mm}
\caption{Annotated sample and its score in Holmes-SFTSet for Qwen2VL-72B.}
\label{fig:chose1}
\vspace{-8mm}
\end{figure*}

\begin{figure*}[h]
\centering
\includegraphics[width=0.8\textwidth]{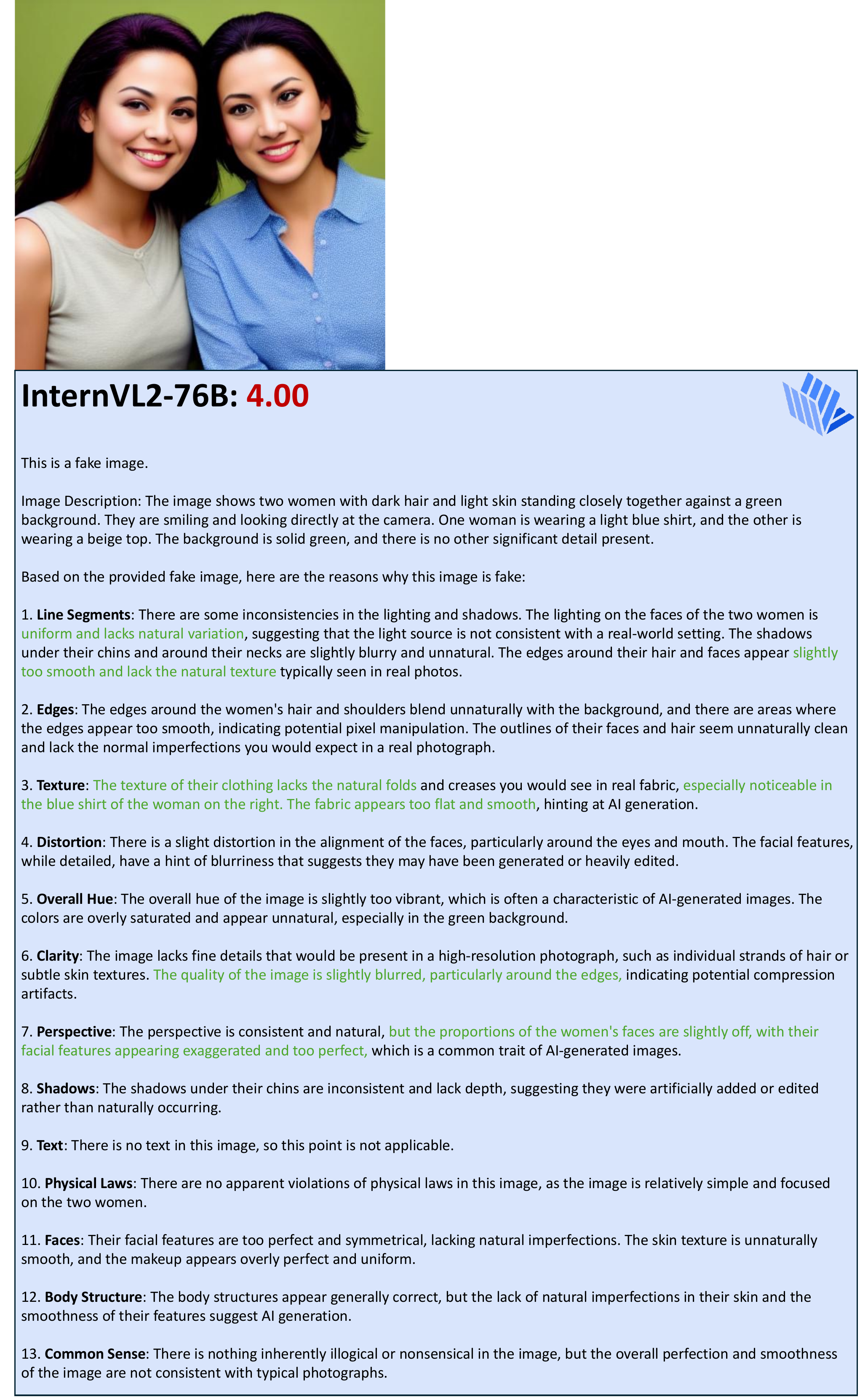}
\vspace{-3mm}
\caption{Annotated sample and its score in Holmes-SFTSet for InternVL2-76B.}
\label{fig:chose2}
\vspace{-8mm}
\end{figure*}

\begin{figure*}[h]
\centering
\includegraphics[width=0.99\textwidth]{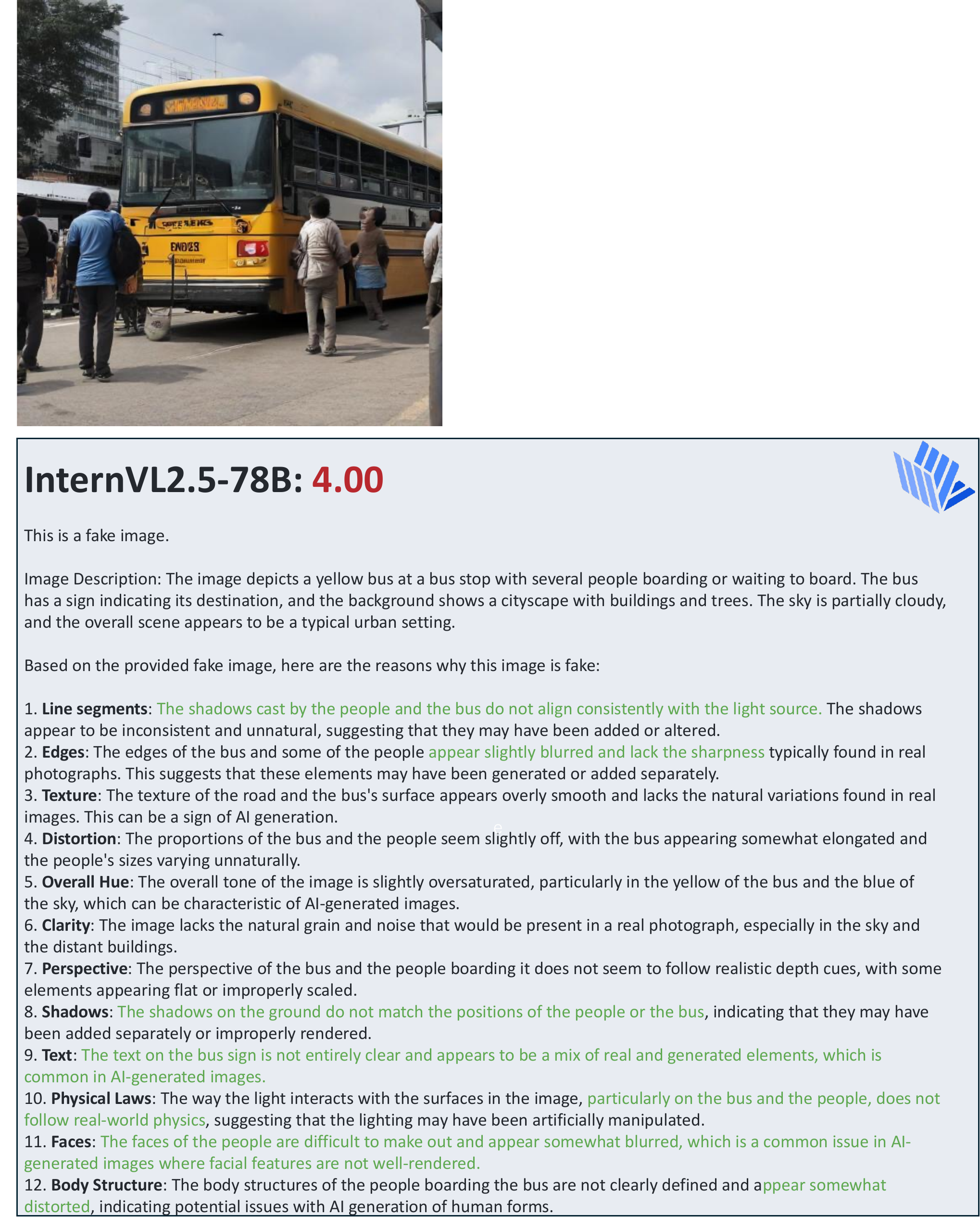}
\vspace{-3mm}
\caption{Annotated sample and its score in Holmes-SFTSet for InternVL2.5-78B.}
\label{fig:chose3}
\vspace{-8mm}
\end{figure*}

\begin{figure*}[h]
\centering
\includegraphics[width=0.99\textwidth]{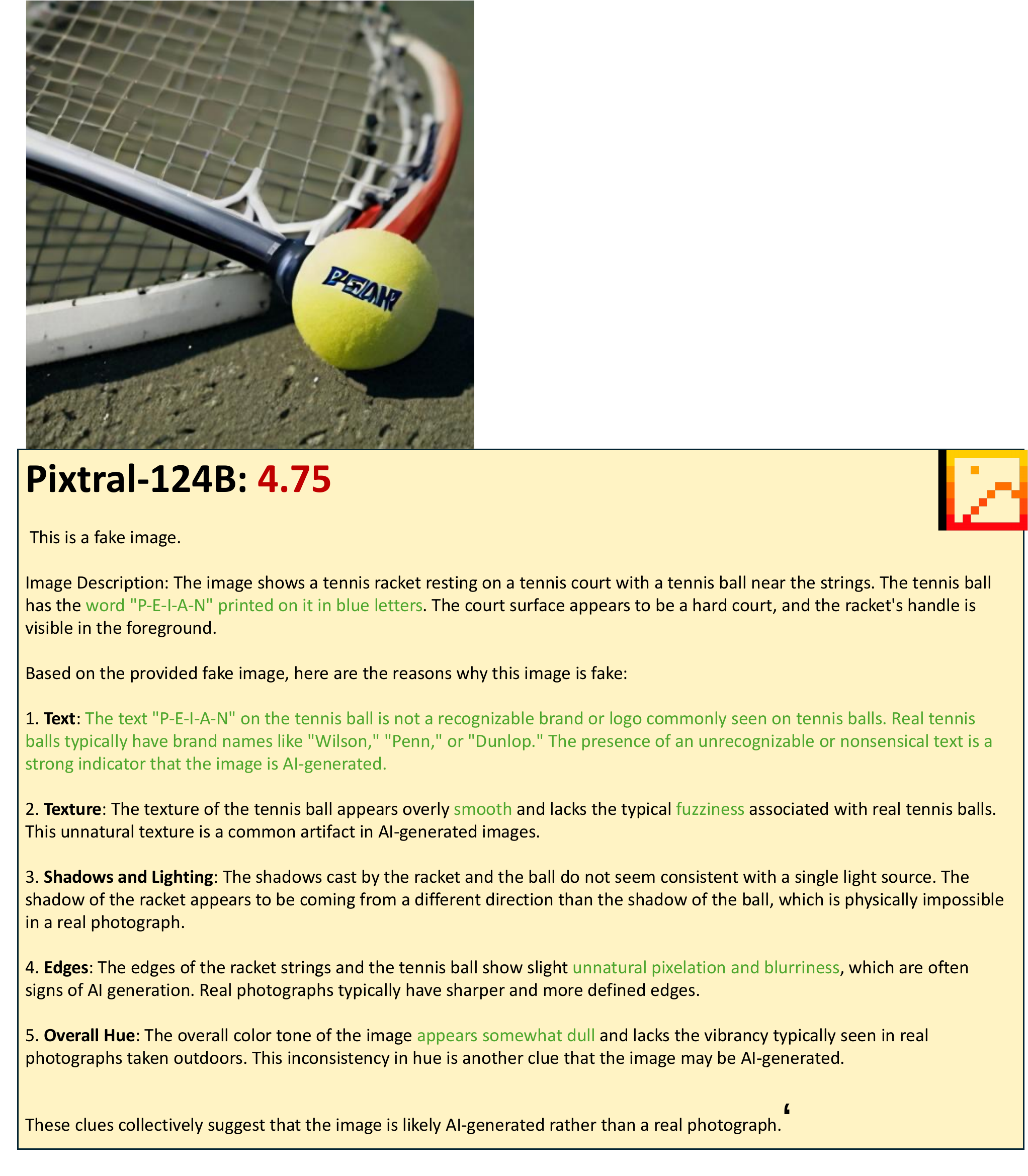}
\vspace{-3mm}
\caption{Annotated sample and its score in Holmes-SFTSet for Pixtral-124B.}
\label{fig:chose4}
\vspace{-8mm}
\end{figure*}

\noindent \textbf{Multi-Expert Jury.} We conducted an analysis of the cross-validation scores of the models. In Fig.~\ref{fig:stat}, the leftmost image shows the distribution of scores given by different models to other models, where {ModelA}\_{ModelB} represents the score given by Model A to Model B. The middle image represents the correlation of scores given to different models, and the rightmost image represents the adoption of annotations from various models in the image-text pairs. From the figures, we can observe that the scoring differentiation of the Pixtral-124B and InternVL-76B models is relatively higher, while the scoring correlation between InternVL2.5-78B and Qwen2VL-72B is very high. This may be due to the use of the same large language model base, Qwen, which could have affected these models' ability to evaluate AI-generated image explanations due to safety alignment operations. Ultimately, our dataset is composed mostly of annotations from Pixtral-124B, with some from InternVL-76B, Qwen2VL-72B, and a small amount from InternVL2.5-78B. We present some examples from our SFT dataset in Fig.~\ref{fig:chose1}, Fig.~\ref{fig:chose2}, Fig.~\ref{fig:chose3}, and Fig.~\ref{fig:chose4}. For InternVL-76B, Qwen2VL-72B, and InternVL2.5-78B, we utilized the vllm framework~\cite{kwon2023efficient}, which is widely used for large model inference, to perform local deployment. For Pixtral-124B, we accessed it via the official website's API. 

\begin{figure*}[h]
\centering
\includegraphics[width=0.8\textwidth]{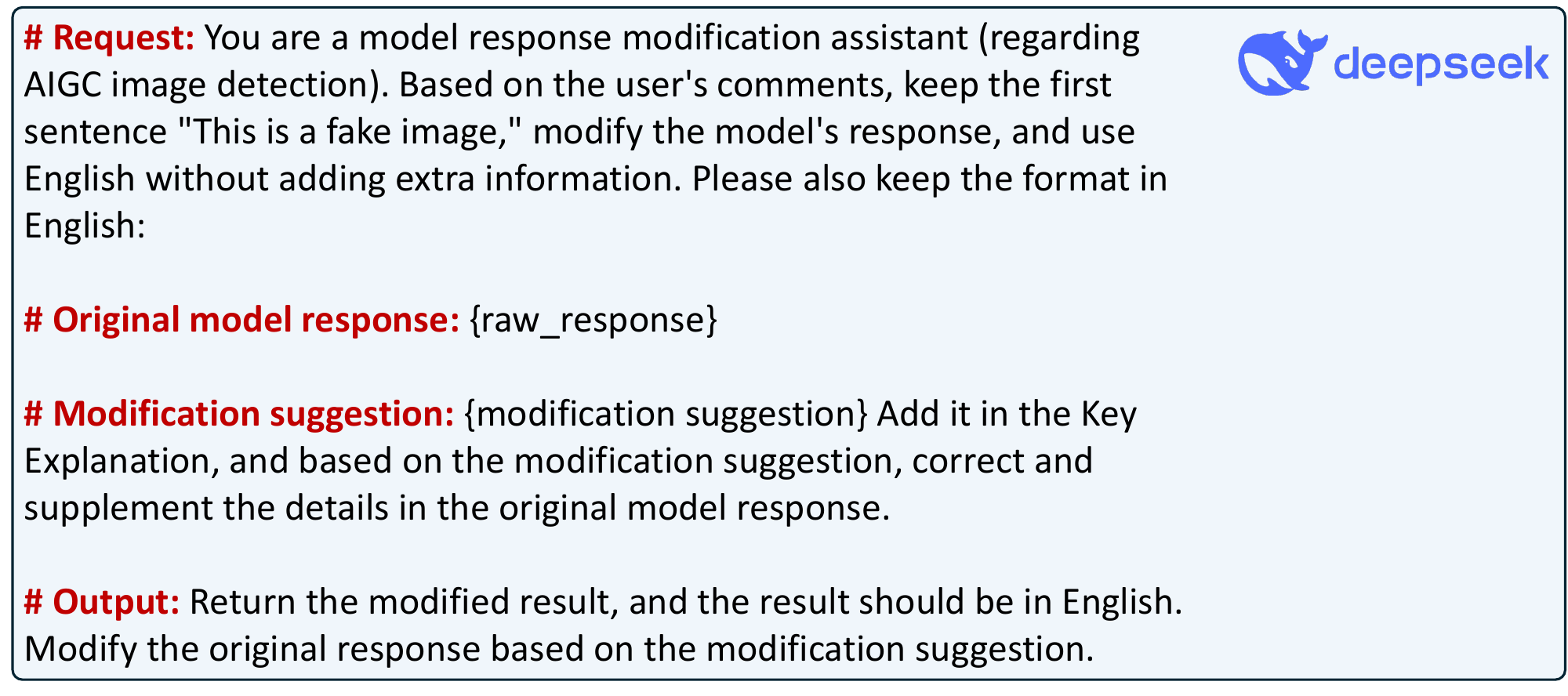}
\vspace{-3mm}
\caption{The original response of the model and the revision suggestions are input into the query prompt of DeepseekV3~\cite{liu2024deepseek}.}
\label{fig:dpsk_prompt}
\vspace{-8mm}
\end{figure*}
\begin{figure*}[h]
\centering
\includegraphics[width=0.99\textwidth]{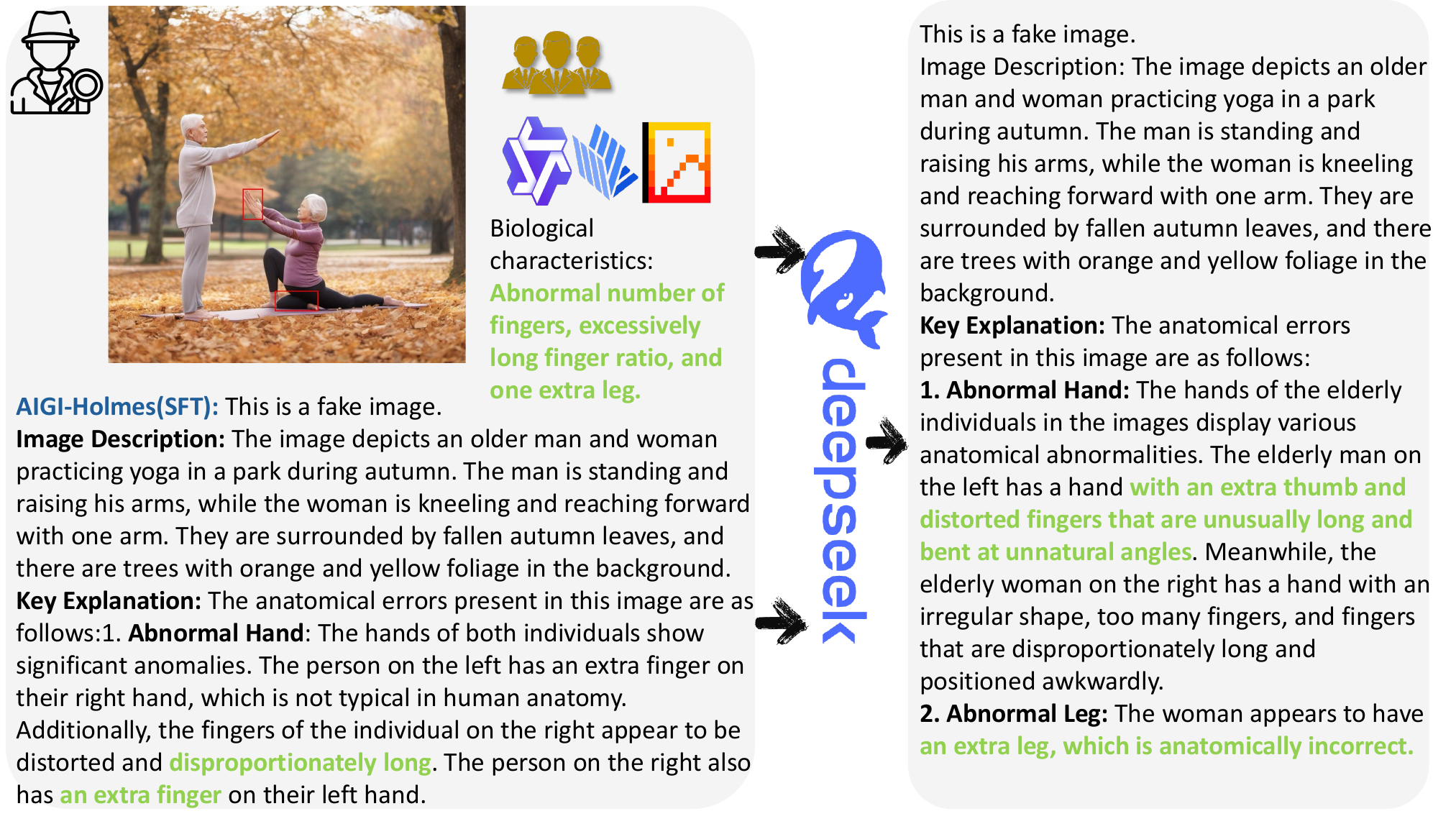}
\vspace{-3mm}
\caption{The schematic flowchart of the DeepseekV3~\cite{liu2024deepseek} generating human-aligned preference samples based on modification suggestions and original responses.}
\label{fig:xiugai}
\vspace{-8mm}
\end{figure*}
\begin{figure*}[h]
\centering
\includegraphics[width=0.83\textwidth]{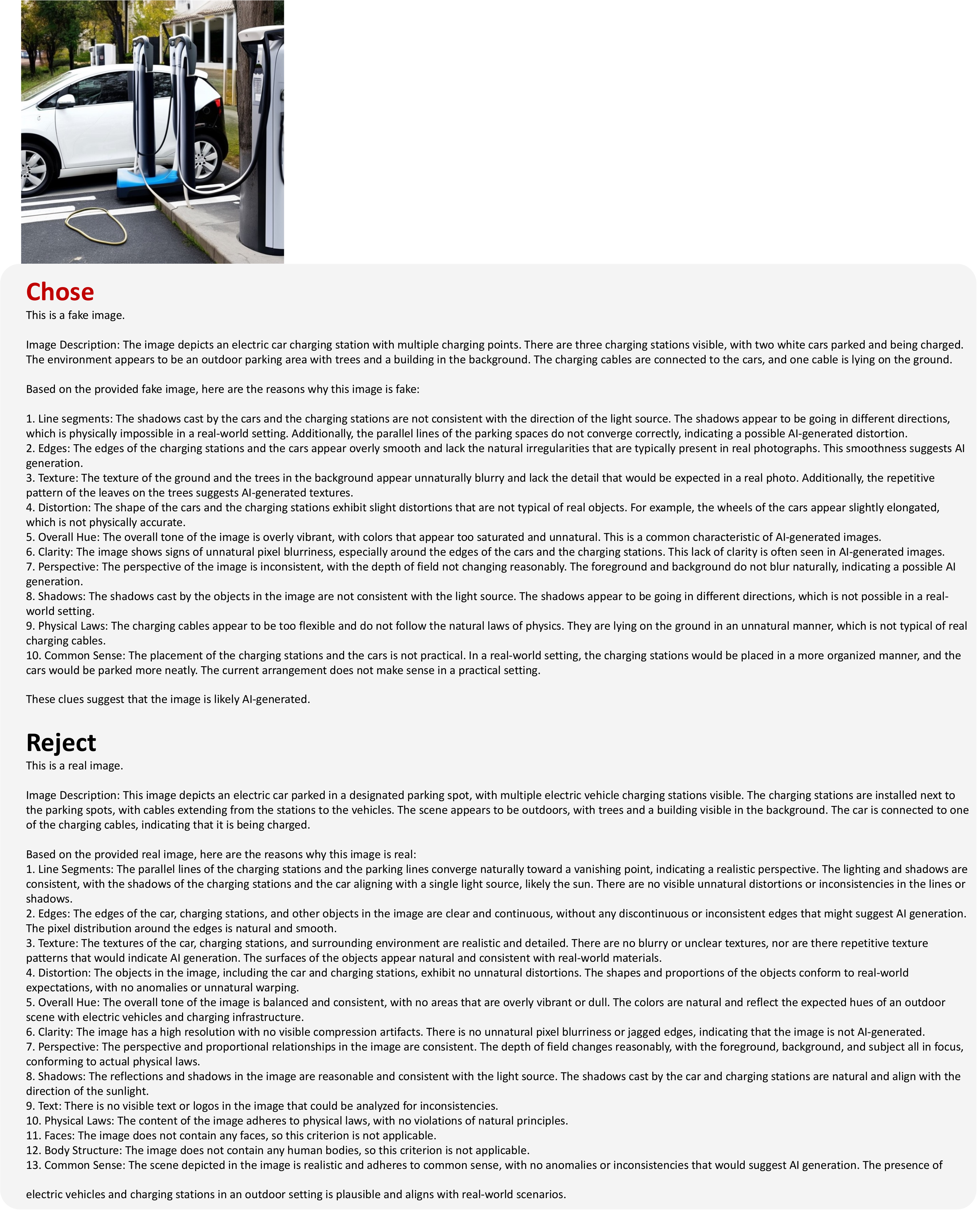}
\vspace{-3mm}
\caption{The presentation of preference sample pairs in $\mathcal{D}_1$.}
\label{fig:dpo1}
\vspace{-8mm}
\end{figure*}
\begin{figure*}[h]
\centering
\includegraphics[width=0.83\textwidth]{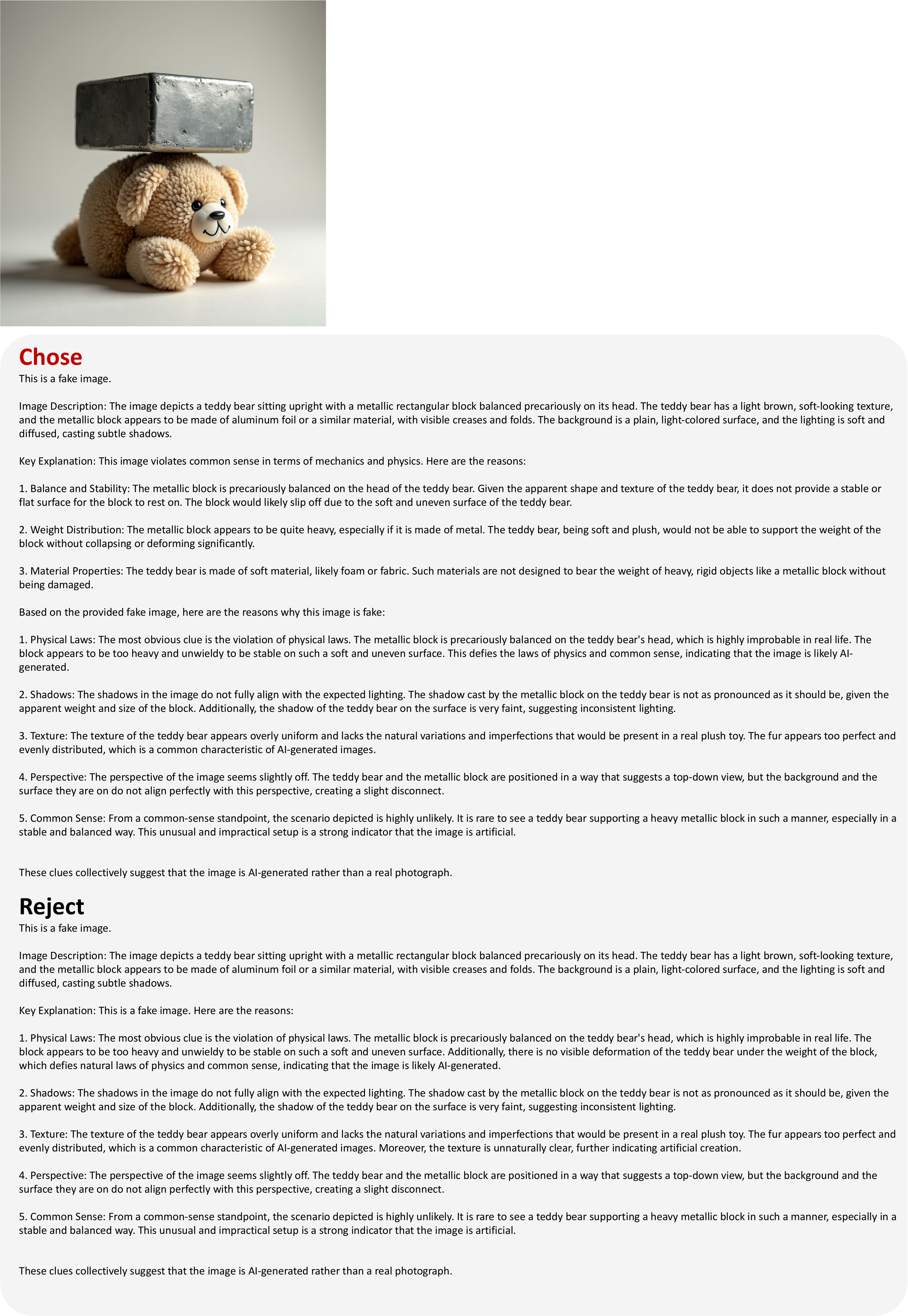}
\vspace{-3mm}
\caption{The presentation of preference sample pairs in $\mathcal{D}_2$.}
\label{fig:dpo2}
\vspace{-8mm}
\end{figure*}

\noindent \textbf{Holmes-DPOSet.} In the main text, we introduced how we solicited revision suggestions for the responses of the supervised fine-tuning model from human experts and multimodal large language model experts. These suggestions were then fed into DeepseekV3~\cite{liu2024deepseek} using the prompt shown in Fig.~\ref{fig:dpsk_prompt} to obtain the final human-aligned preference samples. This process is illustrated in Fig.~\ref{fig:xiugai}. The revision suggestions originated from the annotation platform we developed, as shown in Fig.~\ref{fig:labellm}, based on LabelLLM\footnotemark[2]. By referring to the revision suggestions, the model effectively supplemented the original response by addressing an additional anatomical error that was previously overlooked, while maintaining a high level of consistency before and after the modification. This is beneficial for executing the DPO training phase. In Fig.~\ref{fig:dpo1} and Fig.~\ref{fig:dpo2}, we present sample pairs from $\mathcal{D}_1$ and $\mathcal{D}_2$, respectively.

\begin{figure*}[h]
\centering
\includegraphics[width=0.83\textwidth]{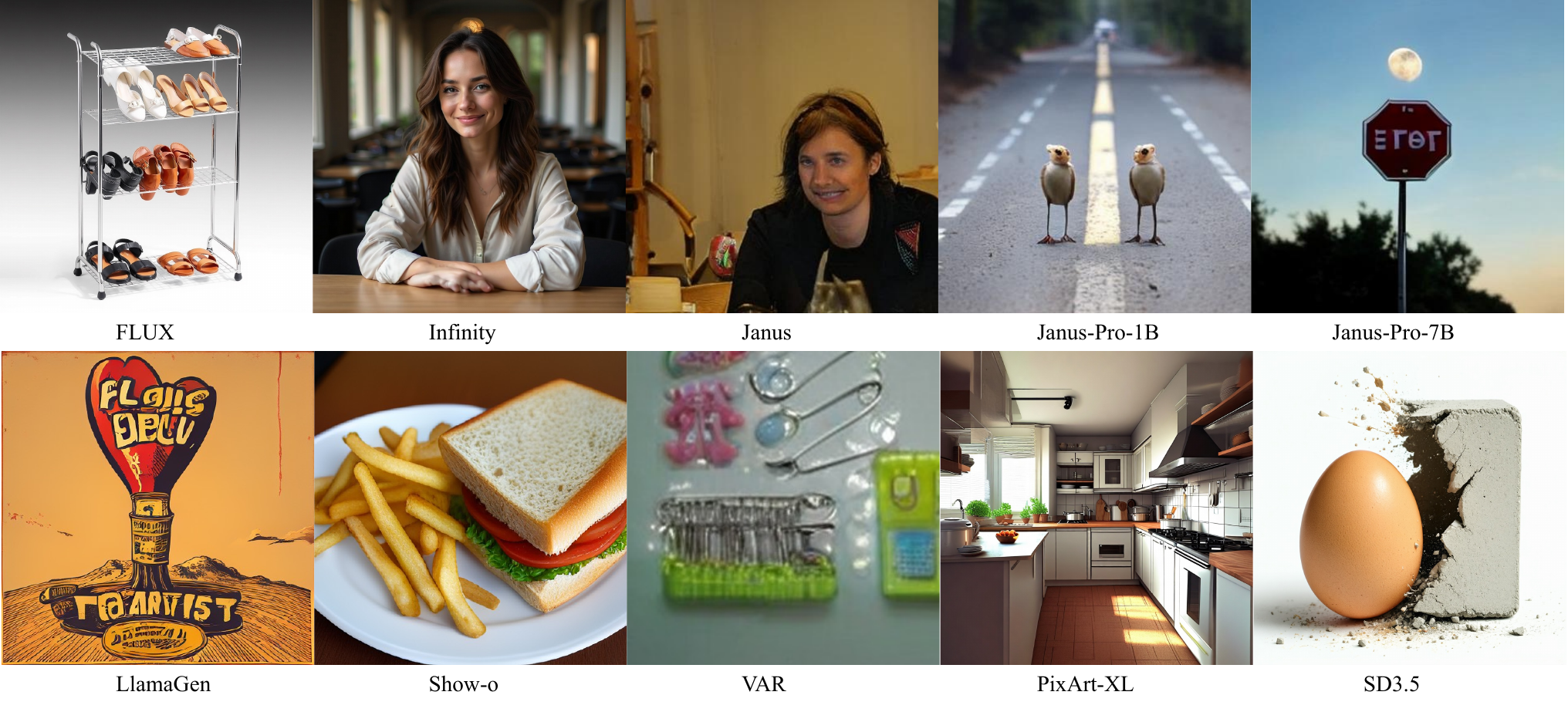}
\vspace{-3mm}
\caption{Qualitative Results of the Test Set for $\mathcal{P}_3$.}
\label{fig:testsample}
\vspace{-8mm}
\end{figure*}

\noindent \textbf{Construction Process of $\mathcal{P}_3$.} For models such as Janus~\cite{wu2024janus}, Janus-Pro~\cite{chen2025janus}, VAR~\cite{VAR}, Infinity~\cite{Infinity}, Show-o~\cite{xie2024showo}, LlamaGen~\cite{sun2024autoregressive}, FLUX~\cite{flux}, and SD3.5~\cite{esser2024scaling}, we deployed these models to perform inference and generate images. The test images for PixArt-XL~\cite{chen2024pixartsigma} and some of the FLUX~\cite{flux} images were sourced from~\cite{li2024improving}. The image resolutions are as follows: FLUX, SD3.5, Infinity, and PixArt-XL have a resolution of $1024 \times 1024$; Show-o and LlamaGen have a resolution of $512 \times 512$; Janus and Janus-Pro have a resolution of $384 \times 384$; and VAR has a resolution of $256 \times 256$. Examples of these images can be found in Fig.~\ref{fig:testsample}.

\footnotetext[1]{\label{fn:1}https://github.com/PaddlePaddle/PaddleOCR}
\footnotetext[2]{\label{fn:2}https://github.com/opendatalab/LabelLLM}




\section{More Experiments}
\label{supp:d}
\subsection{ALL Baselines}
\label{supp:d1}
We compared various baselines including CNNSpot~\cite{wang2020cnn}, FreDect~\cite{frank2020leveraging}, Fusing~\citep{ju2022fusing}, LNP~\citep{liu2022detecting}, LGrad~\citep{tan2023learning}, UnivFD~\citep{ojha2023towards}, DIRE~\citep{wang2023dire}, PatchCraft~\citep{zhong2023rich}, NPR~\citep{tan2024rethinking}, AntiFakePrompt~\cite{chang2023antifakeprompt}, Fatformer~\cite{liu2024forgeryaware}, Ricker2022~\cite{ricker2022towards}, DE-FAKE~\cite{DE-FAKE}, LASTED~\cite{lasted}, QAD~\cite{le2023quality}, InstructBLIP~\cite{dai2023instructblip}, CogVLM~\cite{wang2023cogvlm}, LaRE~\cite{luo2024lare}, RINE~\cite{koutlis2024leveraging}, AIDE~\cite{yan2024sanity}. These baseline methods are trained and tested under fair and consistent conditions across different settings.

\subsection{Training Details of Baselines in Protocol-III}
The training details for the methods we trained on the Holmes-SFTSet training set are as follows, with all unmentioned details consistent with the code provided by the original authors:
\begin{itemize}
\item \textbf{CNNSpot}: We trained using the Adam optimizer with a learning rate of 2e-4 and a batch size of 32 for 10 epochs.
\item \textbf{AntifakePrompt}: We trained using the Adam optimizer with a learning rate of 2e-4 and a batch size of 8 for 10 epochs.
\item \textbf{UnivFD}: We trained using the Adam optimizer with a learning rate of 1e-4 and a batch size of 48 for 20 epochs.
\item \textbf{NPR}: We trained using the Adam optimizer with a learning rate of 2e-4 and a batch size of 32 for 30 epochs.
\item \textbf{LaRE}: We trained using the Adam optimizer with a learning rate of 1e-4 and a batch size of 48 for 10 epochs.
\item \textbf{RINE}: We trained using the Adam optimizer with a learning rate of 2e-4 and a batch size of 16 for 5 epochs.
\item \textbf{AIDE}: We trained using the Adam optimizer with a learning rate of 2e-4 and a batch size of 8 for 5 epochs.
\end{itemize}

\subsection{Comparisons with SOTAs}
\label{supp:d3}
\textbf{Protocol-I.} The quantitative results in Tab.~\ref{table:p1} show the classification accuracy of various methods and generators within the range of $\mathcal{P}_1$. In this evaluation, all methods were trained solely on the four categories (car, cat, chair, horse) generated by ProGAN, except for DIRE-D, which was trained on the Diffusion dataset of ADM. AIGI-Holmes demonstrates significant improvements compared to the current state-of-the-art (SOTA) methods PatchCraft and AIDE, with average accuracy increases of 3.9\% and 0.4\%, respectively. The AIDE method integrates semantic, low-frequency, and high-frequency information through a dual-stream structure, showing remarkable detection performance for some Diffusion methods. However, our method not only performs well in detecting Diffusion methods but also excels in GauGAN and BigGAN methods, where these SOTA methods underperform, with improvements of 10.6\% and 21.8\%, respectively. Although trained only on the single forgery method of ProGAN, which can easily lead to overfitting for mllm, our model still demonstrates good generalization to unseen diffusion methods, highlighting the potential of our approach.

\noindent \textbf{Protocol-II.} The quantitative results in Tab.~\ref{table:p2} show the classification accuracy of various methods and generators within the range of $\mathcal{P}_2$. All methods were trained or fine-tuned on the dataset primarily generated by SD3, highlighted in gray in the table. AIGI-Holmes shows significant improvements compared to the current state-of-the-art (SOTA) method AntifakePrompt, with an average accuracy increase of 1.14\%. The AntifakePrompt method uses prompt learning to perform binary classification on images with mllm, achieving excellent detection performance on a large number of unseen diffusion methods, image inpainting methods, image super-resolution methods, and Deepfake datasets. However, our method achieves over 10\% improvement in detection accuracy on LaMa, DALLE-3, and real image datasets COCO and Flickr. In addition to providing accurate binary classification results, it can also output explanations and reasons corresponding to the predicted results.

\begin{table*}[t]
\begin{center}
\resizebox{0.95\textwidth}{!}
{%
\begin{tabular}{lccccccccccccccccc}
\toprule
Method  & \rot{ProGAN} &\rot{StyleGAN} & \rot{BigGAN} & \rot{CycleGAN} &\rot{StarGAN} &\rot{GauGAN} &\rot{StyleGAN2} &\rot{WFIR} &\rot{ADM} &\rot{Glide} & \rot{{Midjourney}} & \rot{{SD v1.4}} & \rot{{SD v1.5}}& \rot{{VQDM}}& \rot{{Wukong}}& \rot{{DALLE2}} & \rot{\textit{Mean}} \\ \midrule
CNNSpot &\textbf{100.00} &90.17 &71.17 &87.62 &94.60 & 81.42 &86.91 & {91.65} &{60.39} &58.07 & {51.39}&50.57 &50.53 &56.46 &51.03 &50.45 &70.78  \\
FreDect & 99.36 & 78.02 & 81.97 & 78.77 & 94.62 & 80.57 & 66.19 & 50.75 & 63.42 & 54.13 & 45.87 & 38.79 & 39.21 & 77.80 & 40.30 & 34.70 & 64.03 \\
Fusing & \textbf{100.00} & 85.20 & 77.40 & 87.00 & 97.00 & 77.00 & 83.30 & 66.80 & 49.00 & 57.20 & 52.20 & 51.00 & 51.40 & 55.10 & 51.70 & 52.80 & 68.38 \\
LNP & 99.67 & 91.75 & 77.75 & 84.10 & {99.92} & 75.39 & 94.64 & 70.85 & 84.73 & 80.52 & 65.55 & 85.55 & 85.67 & 74.46 & 82.06 & 88.75 & 83.84 \\
LGrad & 99.83 & 91.08 & 85.62 & 86.94 & 99.27 & 78.46 & 85.32 & 55.70 & {67.15} & 66.11 & 65.35 & 63.02 & 63.67 & 72.99 & 59.55 & 65.45 & 75.34 \\
UnivFD  & 99.81 & 84.93 & \underline{95.08} & \underline{98.33} & 95.75 & \textbf{99.47} & 74.96 & 86.90 & {66.87} & 62.46 & 56.13 & 63.66 & 63.49 & 85.31 & 70.93  & 50.75 & 78.43 \\
DIRE-G & 95.19 & 83.03 & 70.12 & 74.19 & {95.47} & {67.79} & 75.31 & 58.05 & 75.78 & 71.75 & 58.01 & 49.74 & 49.83 & 53.68 & 54.46 & 66.48 & 68.68 \\
DIRE-D & 52.75 & 51.31 & 49.70 & 49.58 & 46.72 & 51.23 & 51.72 & 53.30 & \textbf{98.25} & \underline{92.42} & \underline{89.45} & 91.24 & 91.63 & \underline{91.90} & 90.90 & \underline{92.45} & 71.53 \\
PatchCraft & \textbf{100.00} & 92.77 & \textbf{95.80} & {70.17} & \textbf{99.97} & {71.58} & {89.55} & {85.80} & {82.17} & {83.79} & \textbf{{90.12}} & \textbf{95.38 }& \textbf{95.30 }& 88.91 & \underline{91.07} & \textbf{96.60} & \underline{89.31} \\
NPR & 99.79 & {97.70} & 84.35 & 96.10 & 99.35 & {82.50} & \underline{98.38} & 65.80 & 69.69 & 78.36 & 77.85 & 78.63 & 78.89 & 78.13 & 76.11 & 64.90 & 82.91  \\
{{AIDE}} & \underline{99.99} & \textbf{99.64} & {83.95} & \textbf{98.48} & {99.91} & 73.25 & {98.00} & \underline{94.20} & \underline{93.43} & \textbf{{95.09}} & {77.20} & \underline{93.00} & \underline{92.85} & \textbf{95.16} & \textbf{93.55} & \textbf{96.60} & \underline{92.77} \\
\rowcolor{gray!20} 
{\textbf{AIGI-Holmes}} & \textbf{100.00} & \underline{98.35} & 94.51 & 97.03 & \textbf{100.00} & \underline{95.19} & \textbf{98.88} & \textbf{95.71} & 88.43 & 91.53 & 81.56 & 91.28 & 91.38 & 90.94 & 89.46 & 85.32 & \textbf{93.16} \\
\bottomrule
							
\end{tabular}
}
\caption{Evaluation on $\mathcal{P}_1$: All baseline results are trained on $\mathcal{P}_1$'s training set to ensure a fair comparison. The remaining baseline results are sourced from AIDE~\cite{yan2024sanity}.}
\label{table:p1}
\end{center}
\vspace{-4mm}
\end{table*}

\begin{table*}[t]
    \centering


\resizebox{0.95\textwidth}{!}{
    \begin{tabular}{lccccccccccccccccccc}
    \toprule
     & Ricker2022 & ResNet & FatFormer & \multicolumn{2}{c}{CNNSpot} & \multicolumn{2}{c}{DE-FAKE} & \multicolumn{2}{c}{DIRE} & \multicolumn{2}{c}{LASTED} & \multicolumn{2}{c}{QAD} & CogVLM & \multicolumn{2}{c}{InstructBLIP} & \multicolumn{2}{c}{{AntifakePrompt}} & \textbf{Ours} \\ \cmidrule(rl){2-2} \cmidrule(rl){3-3} \cmidrule(rl){4-4} \cmidrule(rl){5-6} \cmidrule(rl){7-8} \cmidrule(rl){9-10} \cmidrule(rl){11-12} \cmidrule(rl){13-14} \cmidrule(rl){15-15} \cmidrule(rl){16-17} \cmidrule(rl){18-19} \cmidrule(rl){20-20}
    \textbf{Dataset} & P & F & P & P & F & P & F & P & F & P & F & P & F & P & P & LoRA & \textbf{Orig.} & \textbf{+LaMa} & \textbf{+LaMa} \\ \midrule
    \textbf{COCO} & 95.60 & \cellcolor[HTML]{D9D9D9}99.43 & 97.40 & 96.87 & \cellcolor[HTML]{D9D9D9}\underline{99.97} & \cellcolor[HTML]{D9D9D9}85.97 & \cellcolor[HTML]{D9D9D9}83.30 & 81.77 & \cellcolor[HTML]{D9D9D9}{99.93} & 75.47 & \cellcolor[HTML]{D9D9D9}58.10 & 59.57 & \cellcolor[HTML]{D9D9D9}96.83 & 98.43 & 98.93 & \cellcolor[HTML]{D9D9D9}97.63 & \cellcolor[HTML]{D9D9D9}92.53 & \cellcolor[HTML]{D9D9D9}90.40 & \cellcolor[HTML]{D9D9D9} \textbf{100.00} \\
    \textbf{Flickr} & 95.80 & 99.23 & 98.13 & 96.67 & \textbf{100.00} & 90.67 & 84.38 & 77.53 & \underline{99.93} & 76.33 & 65.58 & 60.23 & 98.30 & 99.63 & 99.63 & 97.50 & 91.57 & 90.60 & \textbf{100.00} \\ \midrule
    \textbf{SD2} & 81.10 & 2.50 & 16.83 & 0.17 & 5.23 & \cellcolor[HTML]{D9D9D9}97.10 & 88.07 & 3.83 & 30.47 & 58.69 & 52.53 & 51.00 & 10.67 & 52.47 & 40.27 & 89.57 & \textbf{98.33} & \underline{97.97} & 89.61 \\
    \textbf{SD3} & 88.40 & \cellcolor[HTML]{D9D9D9}\underline{99.83} & 21.50 & 4.70 & \cellcolor[HTML]{D9D9D9}8.60 & 96.50 & \cellcolor[HTML]{D9D9D9}95.17 & 0.00 & \cellcolor[HTML]{D9D9D9}98.53 & 78.68 & \cellcolor[HTML]{D9D9D9}79.51 & 46.53 & \cellcolor[HTML]{D9D9D9}\textbf{99.97} & 2.10 & 1.47 & \cellcolor[HTML]{D9D9D9}97.60 & \cellcolor[HTML]{D9D9D9}96.17 & \cellcolor[HTML]{D9D9D9}96.10 & \cellcolor[HTML]{D9D9D9}99.27 \\
    \textbf{SDXL} & 81.10 & 0.50 & 30.39 & 0.17 & 1.53 & 90.50 & 72.17 & 18.17 & 19.73 & 51.33 & 77.65 & 41.60 & 9.87 & 32.57 & 23.07 & 96.47 & \underline{99.17} & \textbf{99.37} & 99.98 \\
    \textbf{IF} & 92.65 & 4.40 & 27.73 & 19.17 & 4.93 & \textbf{99.20} & 95.20 & 6.93 & 63.17 & 57.99 & 55.63 & 59.07 & 15.17 & 29.03 & 20.63 & 87.90 & \underline{97.10} & 95.97 & 96.37 \\
    \textbf{DALLE-2} & 52.10 & 12.80 & 76.03 & 3.40 & 0.87 & 68.97 & 61.17 & 2.13 & 1.50 & 57.96 & 81.91 & 41.70 & 14.63 & 60.70 & 41.77 & \underline{99.27} & 97.27 & {98.00} & \textbf{100.00} \\
    \textbf{DALLE-3} & \underline{95.20} & 2.10 & 43.97 & 18.17 & 3.20 & {89.00} & 71.57 & 0.10 & 36.27 & 51.83 & 53.00 & 51.23 & 9.83 & 6.03 & 6.63 & 67.87 & 80.80 & 82.97 & \textbf{99.68} \\
    \textbf{playground v2.5} & 94.40 & 0.20 & 29.83 & 15.73 & 0.47 & 96.20 & 86.77 & 0.17 & 17.73 & 70.95 & 65.42 & 38.73 & 2.47 & 13.37 & 6.70 & 95.43 & {97.73} & \underline{98.13} & \textbf{98.53} \\
    \textbf{DiffusionDB} & 81.20 & 4.69 & 60.50 & 9.03 & 4.50 & 80.80 & 78.10 & 2.53 & 16.40 & 86.48 & 67.42 & 52.07 & 12.07 & 6.05 & 53.00 & 85.40 & {98.47} & \underline{98.90} & \textbf{99.25} \\
    \textbf{SGXL} & \textbf{100.00} & 1.63 & 97.73 & 79.30 & 2.13 & 56.90 & 50.20 & 45.27 & 9.50 & 64.39 & 65.59 & 46.40 & 4.20 & 60.40 & 69.53 & 91.20 & 99.03 & \underline{99.37} & 89.09 \\
    \textbf{GLIDE} & 83.80 & 49.97 & 79.80 & 17.23 & 5.87 & 76.50 & 50.20 & 4.63 & 41.77 & 54.46 & 68.19 & 53.63 & 50.27 & 59.90 & 37.97 & 92.63 & {98.90} & \underline{99.70} & \textbf{99.81} \\
    \textbf{Stylization} & 75.50 & 0.90 & 85.03 & 11.40 & 4.17 & 63.97 & 55.17 & 9.90 & 6.30 & 50.70 & 67.79 & 51.93 & 7.93 & 42.90 & 33.97 & 82.80 & {94.10} & \underline{95.77} & \textbf{96.03} \\
    \textbf{DF} & 14.20 & 34.20 & 5.10 & 0.30 & 0.03 & 86.97 & 77.17 & 0.27 & 3.77 & 86.38 & 59.36 & {97.43} & 22.73 & 13.80 & 13.83 & 67.43 & 95.03 & \textbf{98.40} & \underline{98.07} \\
    \textbf{DFDC} & 46.90 & 14.20 & 1.60 & 0.00 & 0.00 & 56.13 & 48.57 & 60.13 & 1.03 & 70.19 & 72.42 & 90.40 & 28.50 & 9.00 & 14.07 & 85.47 & \underline{99.83} & \textbf{99.93} & 89.11 \\
    \textbf{FF++} & 20.30 & 37.53 & 71.30 & 5.23 & 0.23 & 78.90 & 70.63 & 25.50 & 31.93 & 70.69 & 56.50 & \cellcolor[HTML]{D9D9D9}\textbf{99.47} & 30.77 & 35.66 & 44.20 & 88.30 & 95.63 & \underline{97.97} & 97.80 \\
    \textbf{LaMa} & 64.30 & 1.87 & {67.03} & 7.53 & 0.07 & 13.03 & 23.00 & 13.23 & 19.47 & 60.53 & \textbf{97.67} & 42.03 & 3.80 & 5.20 & 10.90 & 42.73 & 39.40 & 55.80 & \underline{95.40} \\
    \textbf{SD2IP} & 59.10 & \cellcolor[HTML]{D9D9D9}\underline{99.76} & 85.07 & 1.27 & \cellcolor[HTML]{D9D9D9}7.23 & 16.00 & \cellcolor[HTML]{D9D9D9}75.57 & 11.37 & \cellcolor[HTML]{D9D9D9}86.40 & 56.96 & \cellcolor[HTML]{D9D9D9}\textbf{99.87} & 42.73 & \cellcolor[HTML]{D9D9D9}96.30 & 35.50 & 44.23 & \cellcolor[HTML]{D9D9D9}91.13 & \cellcolor[HTML]{D9D9D9}80.80 & \cellcolor[HTML]{D9D9D9}89.03 & \cellcolor[HTML]{D9D9D9}94.17 \\
    \textbf{LIIF} & 58.90 & 94.43 & 6.60 & 8.30 & 1.07 & 9.73 & 53.67 & 1.10 & 48.77 & 56.46 & 87.34 & 48.07 & 95.83 & 23.47 & \underline{99.93} & 84.63 & 98.50 & \textbf{99.97} & 60.62 \\
    \textbf{SD2SR} & 73.90 & 97.79 & 84.03 & 1.40 & 0.13 & 29.70 & 96.67 & 2.77 & 27.20 & 59.59 & 99.73 & 47.50 & 8.63 & 55.06 & 69.10 & \textbf{99.90} & 99.43 & \underline{99.80} & 99.76 \\
    \textbf{Average} & 72.79 & 37.22 & 51.42 & 18.12 & 10.68 & 68.45 & 70.22 & 16.94 & 36.68 & 64.69 & 70.01 & 55.64 & 33.39 & 37.59 & 41.83 & 84.17 & {91.16} & \underline{93.08} & \textbf{94.22} \\ 
    \bottomrule
    \end{tabular}
}
    \caption{Evaluation on $\mathcal{P}_2$: All baseline results are trained on $\mathcal{P}_2$'s training set to ensure a fair comparison. The remaining baseline results are sourced from AntifakePrompt~\cite{chang2023antifakeprompt}.}
    \label{table:p2}
\end{table*}
\begin{table*}[t]
{
\begin{minipage}[t]{0.32\textwidth}
\centering
\renewcommand\arraystretch{1.}
\setlength{\tabcolsep}{1.0mm}
\resizebox{0.98\textwidth}{!}{
\begin{tabular}{c|c|c}
\toprule[1.5pt]
{LLM} & \textbf{$\mathcal{P}_1$} & \textbf{$\mathcal{P}_3$} \\ \hline
Llama3-8B & 91.5 & 97.8 \\ \hline
Mistral-7B & 93.2 & 99.2 \\ \hline
Vicuna-7B & 93.0 & 98.3 \\ 
\bottomrule[1.5pt]
\end{tabular}
}
\caption{Mean accuracy on $\mathcal{P}_1$ and $\mathcal{P}_3$ for Different LLMs}
\label{tab:llm_acc}
\end{minipage}%
\hspace{0.01\textwidth}%
\begin{minipage}[t]{0.32\textwidth}
\centering
\renewcommand\arraystretch{1.}
\setlength{\tabcolsep}{1.0mm}
\resizebox{0.98\textwidth}{!}{%
\begin{tabular}{c|c|c}
\toprule[1.5pt]
Training Method & \textbf{$\mathcal{P}_1$} & \textbf{$\mathcal{P}_3$} \\
\hline
Lora(only LLM) & 83.3 & 90.1 \\
\hline
ALL Lora & 84.8 & 91.2 \\
\hline
Holmes Pipeline & 86.8 & 98.2 \\
\bottomrule[1.5pt]
\end{tabular}
}
\caption{Mean accuracy on $\mathcal{P}_1$ and $\mathcal{P}_3$ for different training methods.}
\label{tab:train_acc}
\end{minipage}%
\hspace{0.01\textwidth}%
\begin{minipage}[t]{0.32\textwidth}
\centering
\renewcommand\arraystretch{1.}
\setlength{\tabcolsep}{1.0mm}
\resizebox{0.98\textwidth}{!}{
\begin{tabular}{c|c|c|c}
\toprule[1.5pt]
{CLIP} & {NPR} & \textbf{$\mathcal{P}_1$} & \textbf{$\mathcal{P}_3$} \\
\hline
 &  & 86.8 & 98.2 \\
\hline
\checkmark &  & 92.0 & 98.9 \\
\hline
\checkmark & \checkmark & 93.2 & 99.2 \\
\bottomrule[1.5pt]
\end{tabular}
}
\caption{Mean accuracy on $\mathcal{P}_1$ and $\mathcal{P}_3$ for different visual expert}
\label{tab:vis_acc}
\end{minipage}

\vspace{0.5em}

\begin{minipage}[t]{0.65\textwidth}
\centering
\renewcommand\arraystretch{1.}
\setlength{\tabcolsep}{0.8mm}
\resizebox{0.98\textwidth}{!}{
\begin{tabular}{lcccccccc}
\toprule[1.5pt]
Scaling & CNNSpot & Antifakeprompt & UnivFD & NPR & LaRE & RINE & AIDE & AIGI-Holmes \\
\midrule[1pt]
1$\times$ & 72.9 & 83.9 & 83.6 & 84.0 & 85.0 & 96.2 & 97.0 & 99.2 \\
2$\times$ & 73.6 & 84.8 & 83.7 & 84.7 & 85.2 & 96.8 & 97.3 & - \\
4$\times$ & 73.9 & 85.2 & 83.9 & 84.9 & 85.4 & 97.0 & 97.4 & - \\
\bottomrule[1.5pt]
\end{tabular}
}
\caption{Performance comparison across different scaling factors}
\label{tab:scaling_comp}
\end{minipage}%
\hspace{0.02\textwidth}%
\begin{minipage}[t]{0.31\textwidth}
\centering
\renewcommand\arraystretch{1.}
\setlength{\tabcolsep}{1.0mm}
\resizebox{0.98\textwidth}{!}{
\begin{tabular}{lcccc}
\toprule[1.5pt]
\textbf{} & \textbf{LOKI} & \textbf{Chameleon} & $\mathcal{P}_1$ & $\mathcal{P}_2$ \\
\midrule[1pt]
RINE        & 0.790 & 0.562 & 0.960 & 0.903 \\
AIDE        & 0.706 & 0.667 & 0.949 & 0.945 \\
AIGI-Holmes & 0.824 & 0.759 & 0.987 & 0.954 \\
\bottomrule[1.5pt]
\end{tabular}
}
\caption{Cross-dataset evaluation results}
\label{tab:cross_dataset}
\end{minipage}
}
\end{table*}

\subsection{More Ablation Study}
\label{supp:d4}
We primarily conduct ablation experiments on $\mathcal{P}_1$ and $\mathcal{P}_3$ to evaluate the prediction accuracy of the models.

\noindent \textbf{Large language models.} We utilize the mainstream multimodal architecture LLaVA~\cite{liu2024visual} and conduct ablation experiments on this architecture using large language models. Specifically, we experiment with three language models as shown in Tab.~\ref{tab:llm_acc}: Llama3-8B~\cite{llama3modelcard}, Mistral-7B~\cite{Jiang2023Mistral7}, and Vicuna-7B~\cite{zheng2023judging}. The results indicate that the impact on accuracy is minimal when using different large language models, with Mistral-7B showing a slight advantage. Consequently, our final approach employs Mistral-7B as the large language model within the LLaVA architecture.

\noindent \textbf{Training methods.} We conduct ablation experiments on different training approaches to demonstrate the necessity and effectiveness of our Holmes Pipeline. Specifically, we investigate two training strategies: fine-tuning only the large language model using LoRA, and simultaneously applying LoRA training to both the visual component and the large language model. The results, presented in Tab.~\ref{tab:train_acc}, indicate that our Holmes Pipeline effectively adapts the multimodal architecture to the task of AI-generated image detection. Compared to alternative approaches, our method achieves significant improvements in accuracy, with enhancements of 2.0\% and 7.0\% on $\mathcal{P}_1$ and $\mathcal{P}_3$, respectively.

\noindent \textbf{Integration of Visual Experts.} We conduct ablation studies on the visual expert components introduced during collaborative decoding. The experimental results, presented in Tab.~\ref{tab:vis_acc}, demonstrate that both types of visual experts enhance the performance of the method on $\mathcal{P}_1$ and $\mathcal{P}_3$. The improvement on $\mathcal{P}_1$ is more pronounced, with the integration of a single visual expert and all visual experts increasing the detection accuracy by 5.2\% and 6.4\%, respectively. For $\mathcal{P}_3$, the improvements are 0.7\% and 1.0\%, respectively. This discrepancy may be attributed to the fact that $\mathcal{P}_1$ contains only one type of forgery, namely Progan, leading to overfitting in the MLLM and limiting its generalizability. In contrast, the visual experts exhibit better generalization capabilities. Through collaborative decoding, we ensure the generalizability of the model's detection performance.

\noindent \textbf{Impact of Dataset Size.} Due to the high cost of data annotation, we used a relatively small training set ($\sim$ 65K) to train the baselines, which may not be sufficient for the baselines to reach their optimal performance. For a fair comparison, we proportionally increased the size of the training set and retrained the baselines. As shown in Tab.~\ref{tab:scaling_comp}, the results indicate that the marginal benefit of increasing dataset size for the baselines diminishes. In contrast, our method achieves optimal test performance even when trained on only a quarter of the dataset, highlighting the efficiency of our training procedure.

\noindent \textbf{Cross-benchmark Evaluation.} Given the strong performance of our model on $\mathcal{P}_3$, we further evaluate its detection capability across multiple benchmarks to comprehensively demonstrate the effectiveness of our training set and pipeline. Specifically, we compare our model with the two best-performing baselines, RINE~\cite{koutlis2024leveraging} and AIDE~\cite{yan2024sanity}, on LOKI~\cite{ye2024loki}, Chameleon~\cite{yan2024sanity}, $\mathcal{P}_1$, and $\mathcal{P}_2$. On the two challenging benchmarks, LOKI and Chameleon, our model surpasses the second-best model by 3.4\% and 9.2\%, respectively, demonstrating its impressive detection capability.


\subsection{Comparisons with MLLMs}
As shown in the main text, we used the pairwise comparison method from our previous work~\cite{chiang2024chatbot} to compare the responses of different MLLMs. We plotted heatmaps of model comparisons, where the numbers in each heatmap represent the number of times Model A's response was preferred over Model B's response. It can be observed that our model's responses generally achieved better human preference results compared to other models. This demonstrates the effectiveness of our Holmes-DPO. The results in Tab.~\ref{table:exp} in the main text were obtained using the algorithm presented in Algorithm~\ref{algo2}.
\begin{algorithm}
\caption{The method for calculating ELO ratings.}
\begin{algorithmic}[1]
\State \textbf{Initialize:}
\State $\text{r} \gets \text{defaultdict}(\lambda: \text{INIT\_RATING})$
\State $K \gets 4$
\State $\text{SCALE} \gets 400$
\State $\text{BASE} \gets 10$
\State $\text{INIT\_RATING} \gets 1000$

\For{\textbf{each} $key$ \textbf{in} $\text{dic}$}
    \State $\text{model\_a} \gets \text{split}(key, "\_")[0]$
    \State $\text{model\_b} \gets \text{split}(key, "\_")[1]$
    \State $\text{winner} \gets \text{dic}[key]$
    \State $r_a \gets \text{r}[\text{model\_a}]$
    \State $r_b \gets \text{r}[\text{model\_b}]$
    \State $e_a \gets \frac{1}{1 + \text{BASE}^{\frac{r_b - r_a}{\text{SCALE}}}}$
    \State $e_b \gets \frac{1}{1 + \text{BASE}^{\frac{r_a - r_b}{\text{SCALE}}}}$
    
    \If{$\text{winner} = \text{"choice\_A"}$}
        \State $s_a \gets 1$
    \ElsIf{$\text{winner} = \text{"choice\_B"}$}
        \State $s_a \gets 0$
    \ElsIf{$\text{winner} = \text{"choice\_C"}$ \textbf{or} $\text{winner} = \text{None}$}
        \State $s_a \gets 0.5$
    \Else
        \State \textbf{raise} $\text{Exception}(\text{"unexpected vote"} \ \text{winner})$
    \EndIf
    
    \State $\text{r}[\text{model\_a}] \gets \text{r}[\text{model\_a}] + K \cdot (s_a - e_a)$
    \State $\text{r}[\text{model\_b}] \gets \text{r}[\text{model\_b}] + K \cdot (1 - s_a - e_b)$
\EndFor


\end{algorithmic}
\label{algo2}
\end{algorithm}

\section{More Qualitative Results}
\label{supp:e}
\begin{figure*}[h]
\centering
\includegraphics[width=0.9\textwidth]{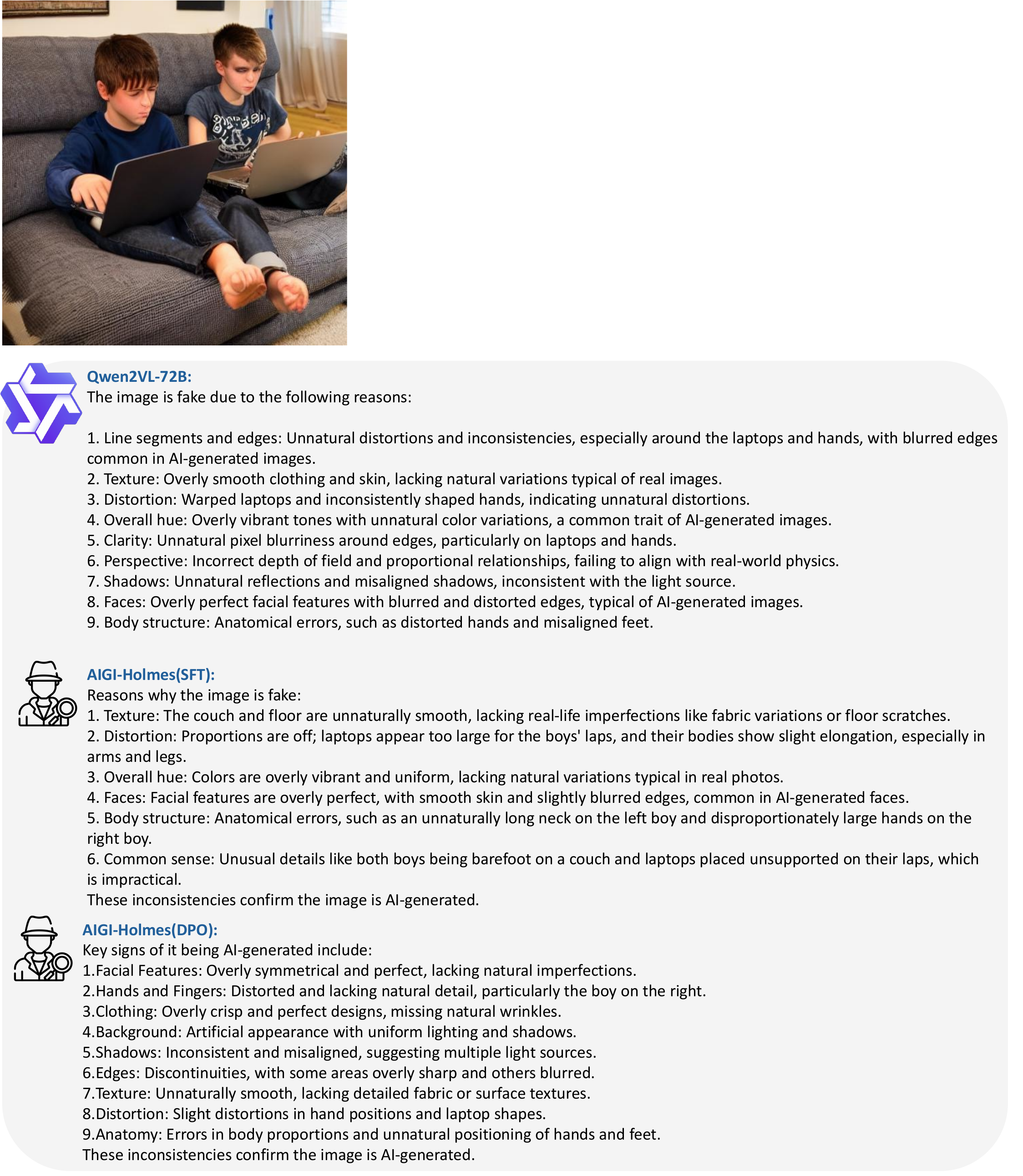}
\vspace{-3mm}
\caption{Comparison of qualitative results between Qwen2VL-72B and our method.}
\label{fig:sample1}
\vspace{-8mm}
\end{figure*}

\begin{figure*}[h]
\centering
\includegraphics[width=0.9\textwidth]{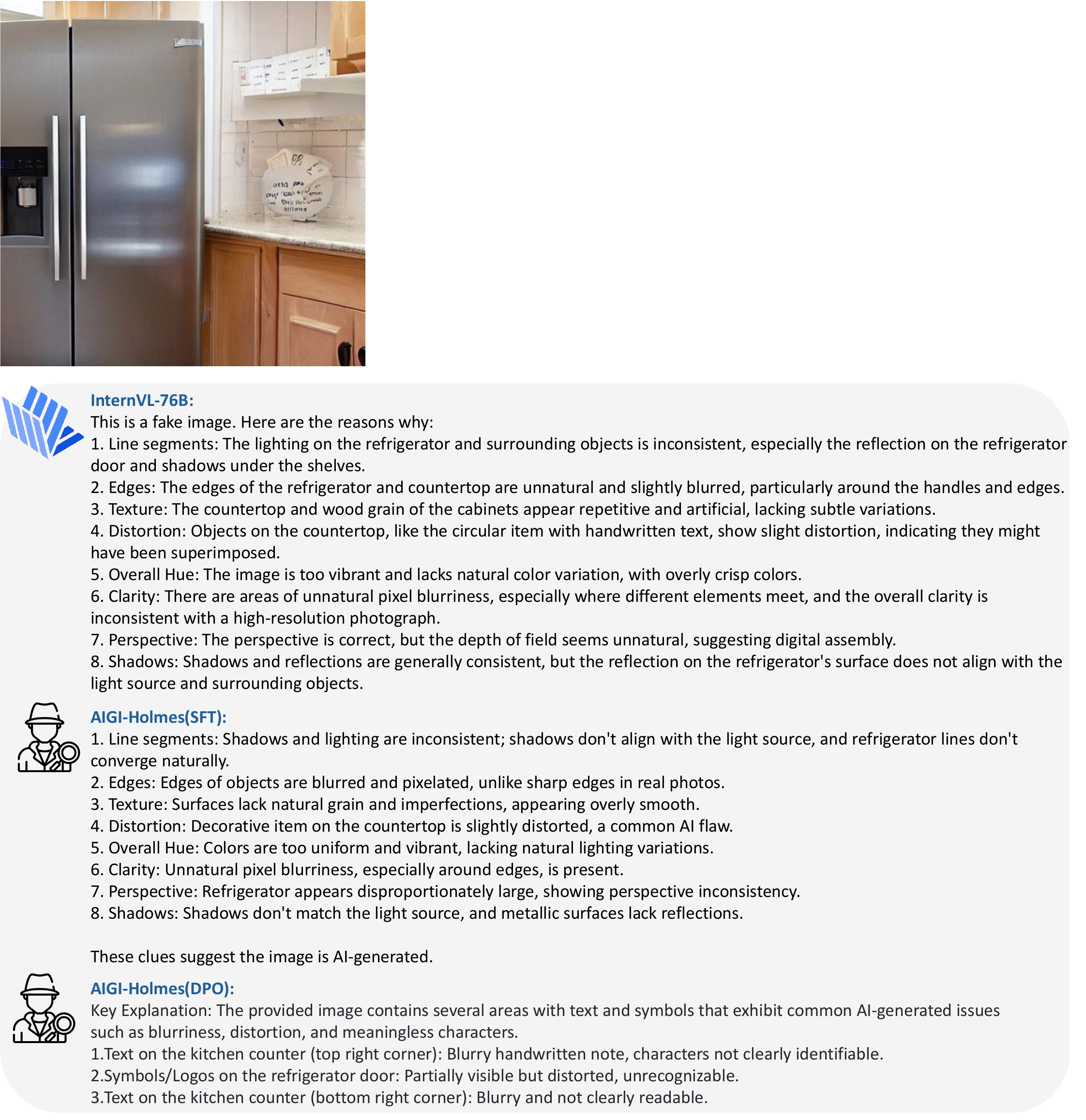}
\vspace{-3mm}
\caption{Comparison of qualitative results between Intern2VL-76B and our method.}
\label{fig:sample2}
\vspace{-8mm}
\end{figure*}

\begin{figure*}[h]
\centering
\includegraphics[width=0.9\textwidth]{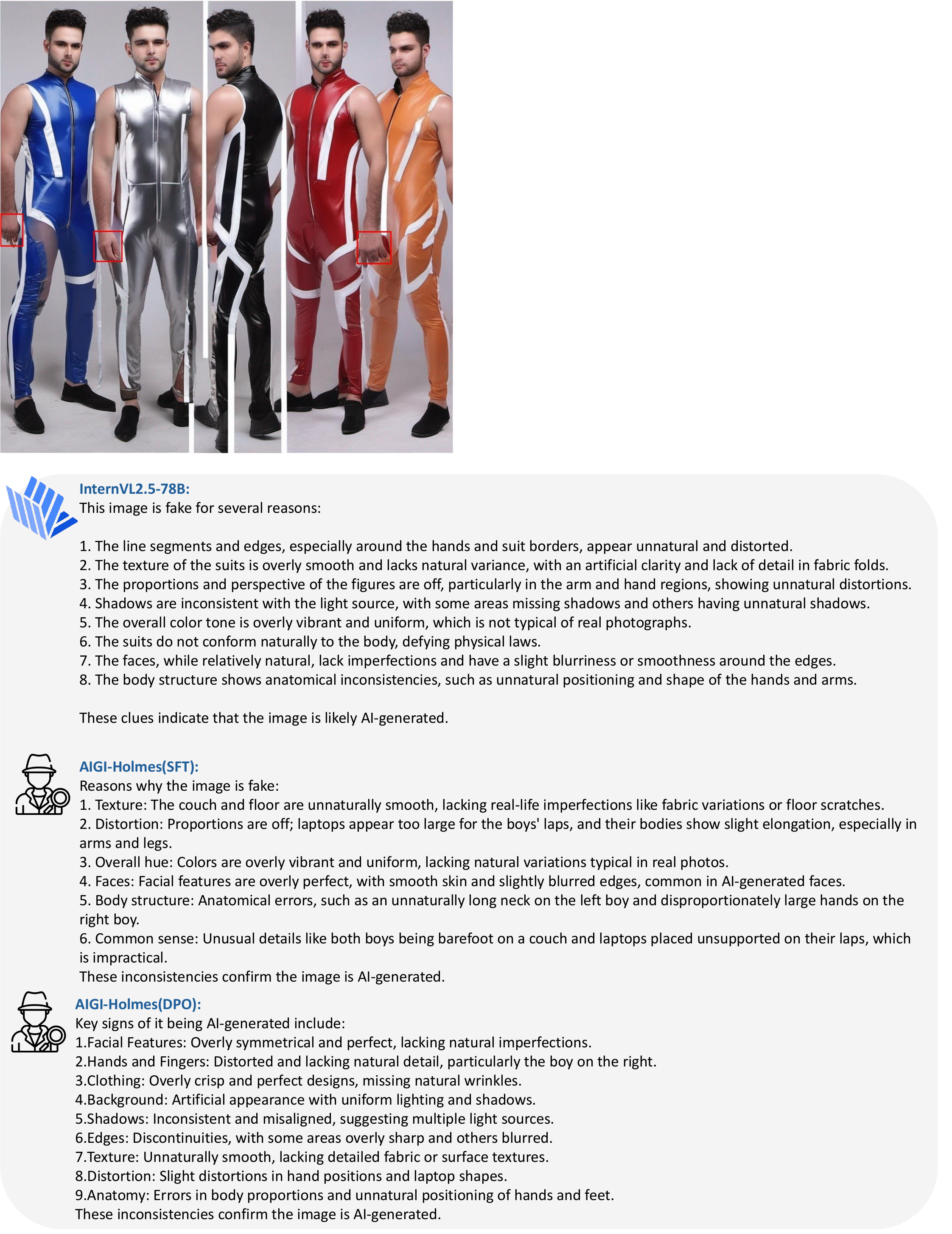}
\vspace{-3mm}
\caption{Comparison of qualitative results between InternVL2.5-78B and our method.}
\label{fig:sample3}
\vspace{-8mm}
\end{figure*}

\begin{figure*}[h]
\centering
\includegraphics[width=0.9\textwidth]{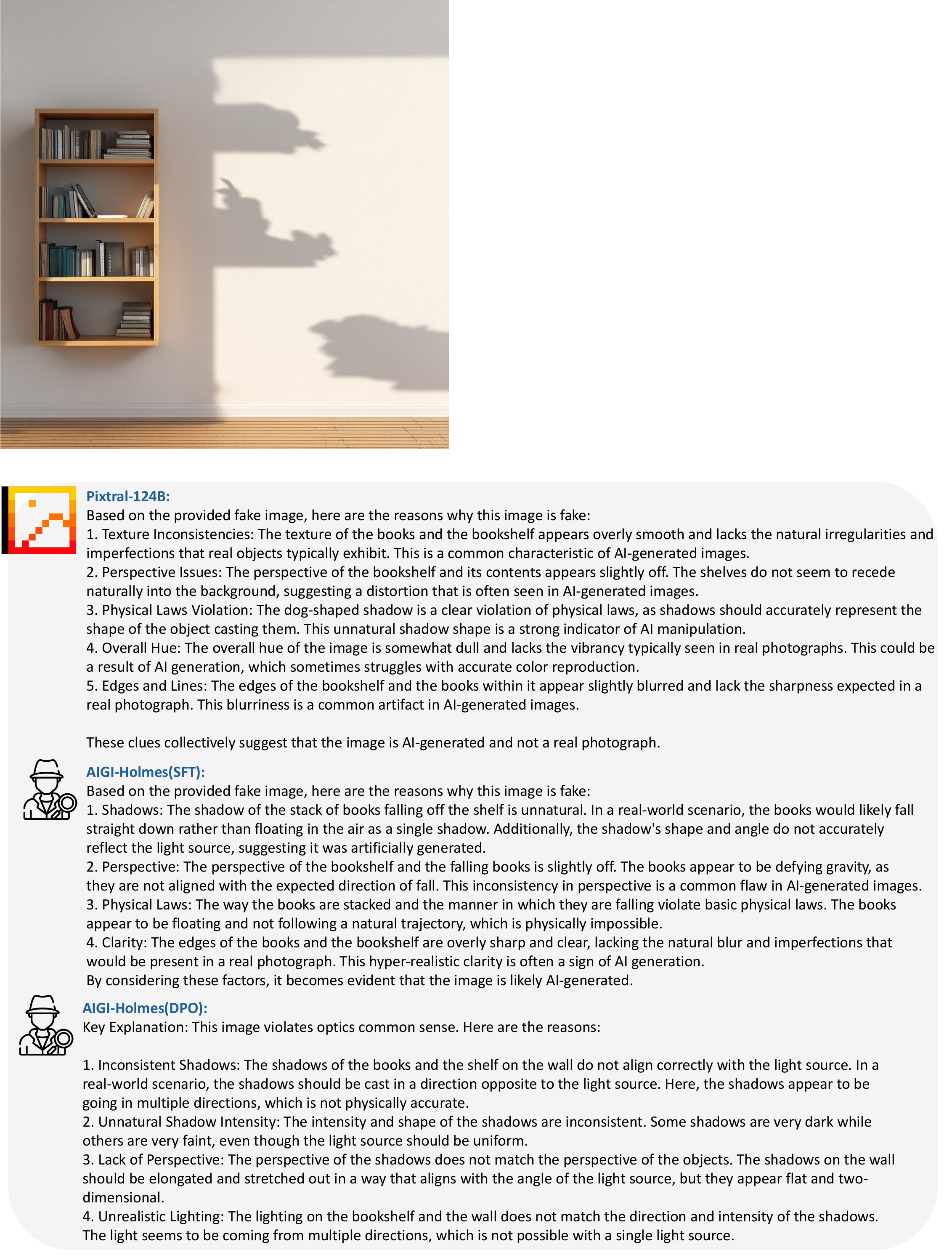}
\vspace{-3mm}
\caption{Comparison of qualitative results between Pixtral-124B and our method.}
\label{fig:sample4}
\vspace{-8mm}
\end{figure*}

In Fig.~\ref{fig:sample1}, Fig.~\ref{fig:sample2}, Fig.~\ref{fig:sample3}, and Fig.~\ref{fig:sample4}, we present a comparison of the explanations provided by the baseline MLLMs method discussed in the main text. Additionally, we illustrate the differences between the SFT-tuned model AIGI-Holmes (SFT) and the DPO-tuned model AIGI-Holmes (DPO). The AIGI-Holmes (DPO) model demonstrates a higher quality of responses.

\section{Limitations and Future Works}
We acknowledge two key limitations. First, as generative models rapidly evolve, the types of forgery-related errors may change, potentially reducing the relevance of our current explanation categories. Our proposed method serves as a foundational approach, and we are committed to extending it to accommodate these emerging error types through adaptable interpretative methods. Second, constrained by the current dataset organization, AIGI-Holmes is limited to generating forensic reports and lacks the image-text dialogue capabilities inherent in multimodal large language models. This limitation of report-only output is also noted in related works \cite{huang2024ffaa, chen2024textitx2dfdframeworkexplainableextendable, xu2024fakeshield, sun2024forgerysleuth}. Future work will focus on three aspects to address these limitations: (1) Continuously deploying AIGI-Holmes in real-world scenarios to build larger-scale SFT and DPO datasets, enhancing both robustness against evolving forgeries and explanatory capabilities; (2) Unifying data from Image Forgery Detection and DeepFake Detection within AIGI-Holmes to develop a comprehensive image authenticity detection model; (3) Expanding the dataset to a dialogue format, potentially via specialized tokens to isolate capabilities, thereby equipping the model with multimodal dialogue functionality.


\begin{figure*}[h]
\centering
\includegraphics[width=0.99\textwidth]{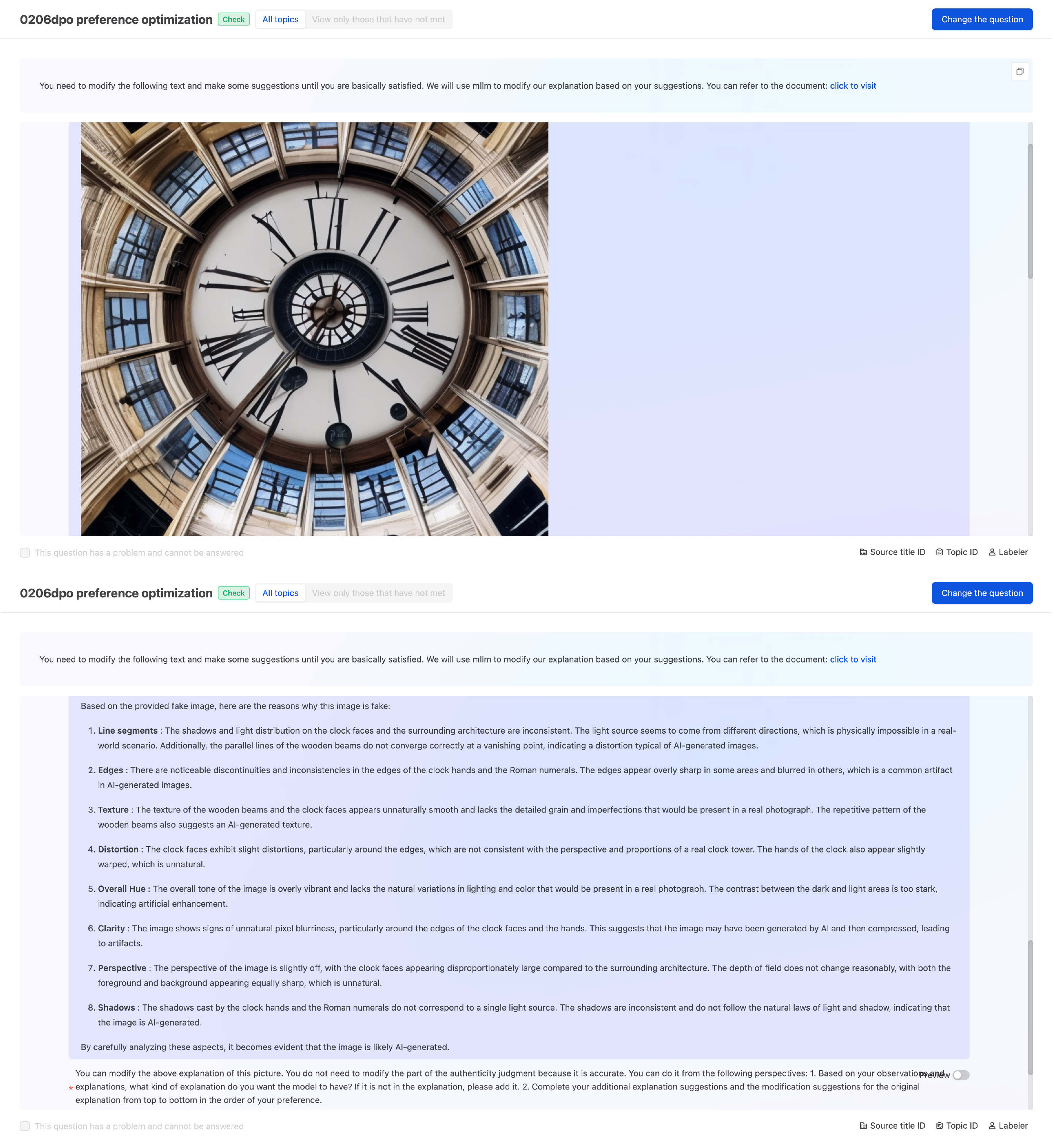}
\vspace{-3mm}
\caption{The annotation interface for obtaining preference samples with modifications suggested by human experts.}
\label{fig:labellm}
\vspace{-8mm}
\end{figure*}

\begin{figure*}[h]
\centering
\includegraphics[width=0.99\textwidth]{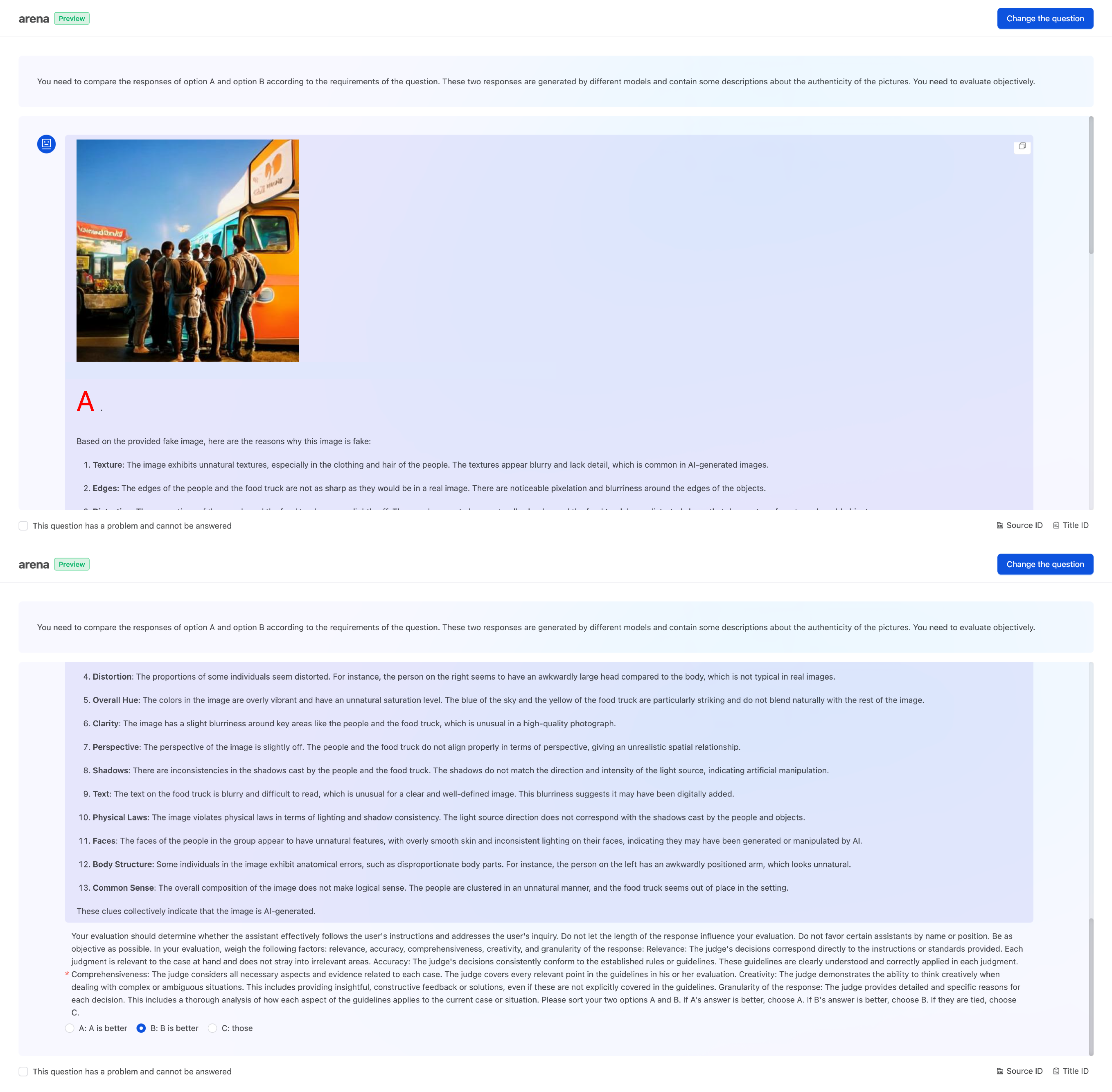}
\vspace{-3mm}
\caption{The interface for evaluating sample selection of human preferences in Arenas.}
\label{fig:labellm}
\vspace{-8mm}
\end{figure*}

\begin{figure*}[h]
\centering
\includegraphics[width=0.99\textwidth]{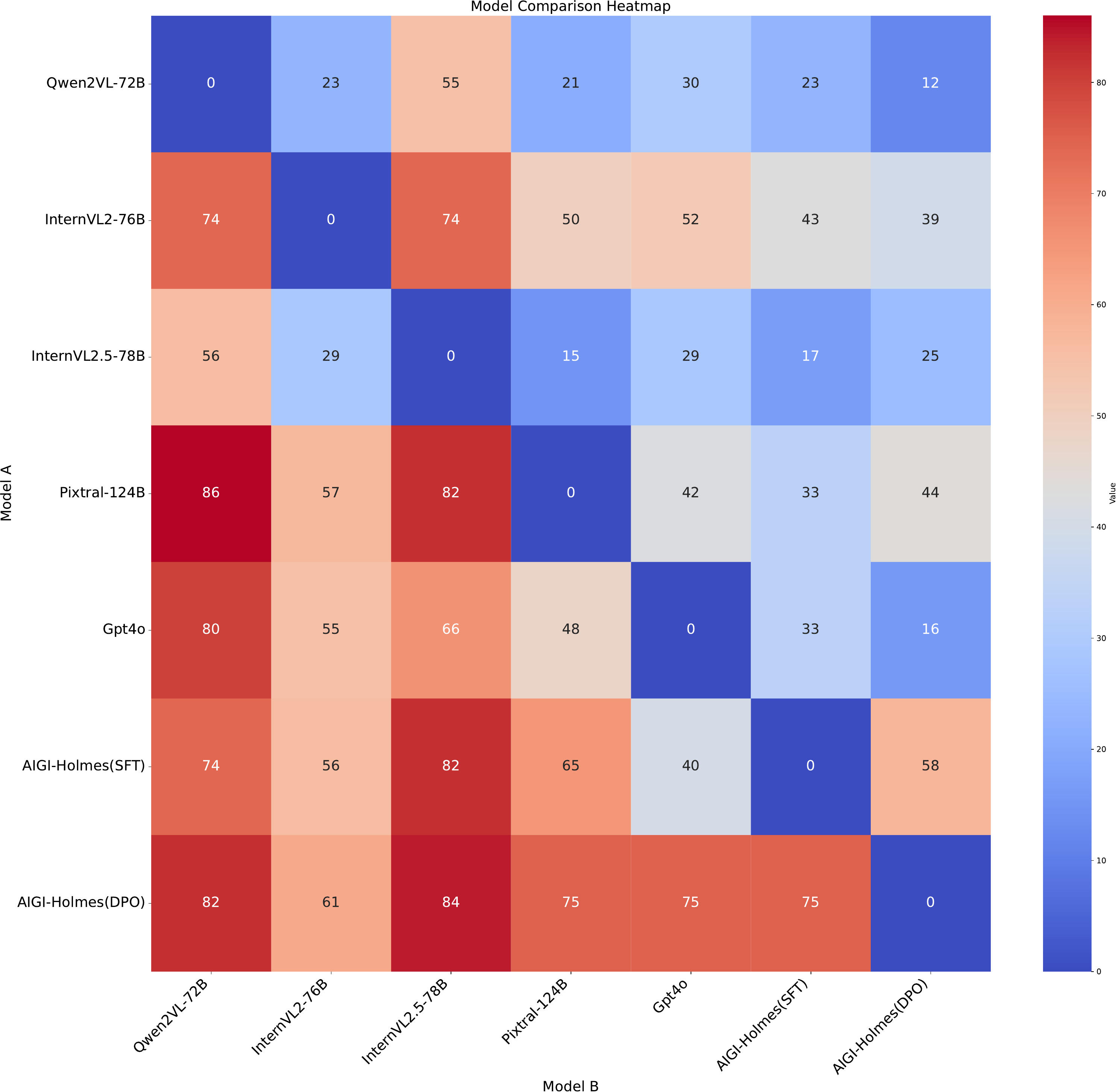}
\vspace{-3mm}
\caption{A heatmap of the winning counts in pairwise scoring among state-of-the-art multimodal language models (MLLMs)}
\label{fig:labellm}
\vspace{-8mm}
\end{figure*}

\end{document}